%% file: main.tex
\documentclass[letterpaper]{article}
\usepackage{aaai2026}
\usepackage{times}
\usepackage{helvet}
\usepackage{courier}
\usepackage[hyphens]{url}
\usepackage{graphicx}
\usepackage{natbib}
\usepackage{caption}
\usepackage{algorithm}
\usepackage{algorithmic}

\usepackage{newfloat}
\usepackage{listings}
\DeclareCaptionStyle{ruled}{labelfont=normalfont,labelsep=colon,strut=off}
\floatstyle{ruled}
\newfloat{listing}{tb}{lst}{}
\floatname{listing}{Listing}

\usepackage{booktabs}
\usepackage{pifont}
\usepackage{amssymb}
\usepackage{amsmath}
\usepackage{enumitem}
\usepackage{bm}
\usepackage[table]{xcolor}

\newcommand{\shline}{\specialrule{0.15em}{0em}{0em}}

\newcommand{\methodname}{SegDINO3D}
\title{{\methodname}: 3D Instance Segmentation Empowered by \\ Both Image-Level and Object-Level 2D Features}

\author {
    Jinyuan Qu\textsuperscript{\rm 1,5}\equalcontrib,
    Hongyang Li\textsuperscript{\rm 2,5}\equalcontrib,
    Xingyu Chen\textsuperscript{\rm 3,5},
    Shilong Liu\textsuperscript{\rm 4},\\
    Yukai Shi\textsuperscript{\rm 1,5},
    Tianhe Ren\textsuperscript{\rm 5},
    Ruitao Jing\textsuperscript{\rm 1,5},
    Lei Zhang\textsuperscript{\rm 5}\thanks{Corresponding author.}
}
\affiliations {
    \textsuperscript{\rm 1}Tsinghua University\\
    \textsuperscript{\rm 2}South China University of Technology\\
    \textsuperscript{\rm 3}Zhongguancun Academy\\
    \textsuperscript{\rm 4}Princeton University\\
    \textsuperscript{\rm 5}International Digital Economy Academy (IDEA)\\

}

\newcommand{\queryca}{DACA-2D}

\nocopyright
\begin{document}

\maketitle

\begin{abstract}
In this paper, we present {\methodname}, a novel Transformer encoder-decoder framework for 3D instance segmentation. 
As 3D training data is generally not as sufficient as 2D training images, {\methodname} is designed to fully leverage 2D representation from a pre-trained 2D detection model, including both image-level and object-level features, for improving 3D representation. 
{\methodname} takes both a point cloud and its associated 2D images as input. 
In the encoder stage, it first enriches each 3D point by retrieving 2D image features from its corresponding image views and then leverages a 3D encoder for 3D context fusion. 
In the decoder stage, it formulates 3D object queries as 3D anchor boxes and performs cross-attention from 3D queries to 2D object queries obtained from 2D images using the 2D detection model. 
These 2D object queries serve as a compact object-level representation of 2D images, effectively avoiding the challenge of keeping thousands of image feature maps in the memory while faithfully preserving the knowledge of the pre-trained 2D model.
The introducing of 3D box queries also enables the model to modulate cross-attention using the predicted boxes for more precise querying.
{\methodname} achieves the state-of-the-art performance on the ScanNetV2 and ScanNet200 3D instance segmentation benchmarks.
Notably, on the challenging ScanNet200 dataset, {\methodname} significantly outperforms prior methods by \textbf{+8.6} and \textbf{+6.8} mAP on the validation and hidden test sets, respectively, demonstrating its superiority.
\end{abstract}

\begin{links}
    \link{Code}{https://github.com/IDEA-Research/SegDINO3D}
\end{links}

\section{Introduction}
\label{sec:intro}

\begin{figure}[t]
    \centering
        \includegraphics[width=0.98\linewidth]{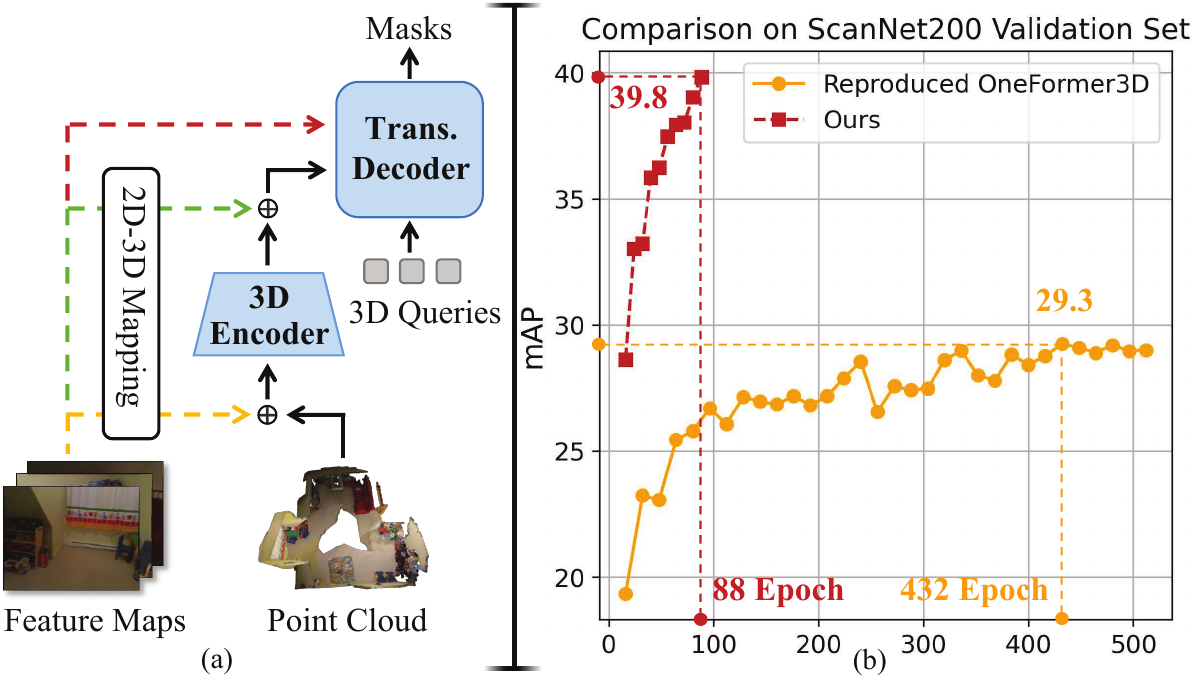}
    \caption{
        (a) The dashed lines indicate different ways of utilizing 2D image features. 
    ``2D-3D Mapping'' denotes the association between 3D points and 2D features, and ``$\oplus$'' indicates feature fusion. 
    (b) {\methodname} obtains significantly better performance and faster convergence speed.
    } 
    \label{fig.teaser}
\end{figure}

3D environment understanding is a crucial capability for AI systems to interact with the physical world. 
While 3D reconstruction has seen remarkable progress in the past few decades from conventional multi-stage geometry-based pipelines~\cite{mur2017orb, schoenberger2016mvs} to more recent end-to-end dense reconstruction approaches~\cite{dust3r, vggt}, 3D semantic understanding, such as 3D object instance detection and segmentation, still remains a challenge, which greatly hinders the development for downstream tasks like manipulation~\cite{gu2024conceptgraphs} and autonomous navigation~\cite{xie2021unseen}.

3D instance segmentation typically takes a 3D point cloud as input and predicts a set of 3D instance masks. 
Recently, motivated by the success of DEtection TRansformers (DETR)~\cite{detr} in 2D detection tasks, Transformer-based frameworks~\cite{Mask3D, spformer} have increasingly dominated the field of 3D instance segmentation. 
They typically follow an encoder-decoder architecture, where a 3D encoder processes the input point cloud and a decoder predicts the object instances.

Despite having achieved great progress, most previous methods (black lines in Fig.\ref{fig.teaser}(a)) only consider point clouds as input, overlooking the accompanying 2D images used for reconstructing the point clouds.
However, benefiting from the abundant 2D data, 2D features trained on the massive 2D images can provide much richer and more reliable semantic information compared to 3D features trained on 3D point clouds.
Therefore, recent works in open-vocabulary scene understanding explore how to leverage 2D foundation models for 3D perception tasks. 
For example, SAM3D~\cite{SAM3D} and SAI3D~\cite{SAI3D} directly use instance proposals generated by 2D perception models but overlook the semantic-rich 2D features.
OpenScene~\cite{openscene} introduces 2D foundation models to extract independent 2D image features (green lines in Fig.~\ref {fig.teaser}(a)), yet lacks global 3D context fusion.
Locate3D~\cite{locate3d} thereby proposes to further train a 3D encoder through self-supervised learning, enhancing the point cloud with both 2D features and global context (yellow lines in Fig.\ref{fig.teaser}(a)).

However, training such a 3D encoder still requires a sufficiently large-scale 3D data set, which is far from comparable with the scale of 2D data available. 
As a result, the 2D features lifted to the 3D space are inadvertently degraded after passing through the 3D encoder and lead to suboptimal performance in 3D instance segmentation.
Motivated by this, we hypothesize that integrating globally-enhanced 2D features in the encoder stage, while simultaneously preserving their semantic discrimination ability in the decoder stage, could yield significant improvements.
Compared to the approach that maps raw 2D image features to discrete 3D points (green lines in Fig.\ref{fig.teaser}(a)), directly attending to 2D image feature maps to update 3D queries (red lines in Fig.\ref{fig.teaser}(a)), as done in BEVFormer~\cite{li2022bevformer}, is generally more effective.
However, for generic 3D instance segmentation, a single scene often contains thousands of images, making it computationally expensive to cross-attend over all of them.

To more effectively leverage 2D image features, we develop {\methodname}, a novel DETR-like 3D instance segmentation framework.
In the encoder stage, to enhance 3D point features with semantically discriminative 2D information, we retrieve and aggregate 2D image-level features for each 3D point from suitable viewpoints through our Nearest View Sampling strategy.
The decorated point clouds are then processed by a 3D encoder for global context fusion, effectively fusing 3D geometric structures with 2D semantic cues.
As the 2D semantic features might be deteriorated after the 3D encoder due to limited 3D training data, we propose to reuse 2D features again in the decoder. 
More specifically, in the decoder stage, we propose a Distance-Aware Cross-Attention ({\queryca}) module.
This module utilizes 2D object queries, instead of 2D image feature maps, from a 2D DETR-based detection model as a compact representation. This effectively addresses the challenge of keeping thousands of image feature maps in memory and attending to them.
As these 2D object queries are directly provided by the well-trained 2D model, they retain strong semantic discrimination ability learned from large-scale 2D data. 
To make {\queryca} more effective, {\queryca} employs a carefully designed distance-aware attention mask to impose spatial constraints, allowing each 3D object query to cross-attend to the most relevant 2D object queries.
Thanks to the symmetric design in dual-level 2D feature enhancement on both 3D points (encoder stage) and object queries (decoder stage), their features are not only enriched but also aligned in a shared space, facilitating more accurate mask decoding via point-query similarity.

In addition, relying solely on the similarity of semantic features, as in previous methods~\cite{Mask3D, spformer}, to query features for 3D object queries is inherently limited. 
Although some of the previous methods try to introduce positional embedding to provide spatial constraints~\cite{maft}, it is still isotropic and insufficient for capturing object scale variations, which are prevalent in 3D environments.
Inspired by 2D DETR-like models~\cite{liu2022dab,zhang2022dino}, we formulate 3D object queries as axis-aligned 3D boxes and introduce an auxiliary regression task. The predicted box sizes can modulate the positional attention map, enabling the 3D object queries to adjust their attention according to object size dynamically.

As shown in Fig.\ref{fig.teaser}(b), {\methodname} not only outperforms previous methods by a large margin on the challenging ScanNet200~\cite{scannet200} dataset, but also achieves significantly faster training convergence.

In summary, our contributions are threefold:
\begin{itemize}[noitemsep, topsep=0.5pt]
    \item We propose {\methodname}, a novel DETR-like encoder-decoder framework for 3D instance segmentation, which effectively utilizes both image-level features and object-level queries from a 2D detection model to improve 3D representations, leading to superior performance.
    \item We introduce 3D box queries to 3D segmentation models and use them to modulate the positional attention map, further enhancing the precision of perception ability.
    \item {\methodname} achieves the state-of-the-art performance on the 3D instance segmentation benchmarks ScanNet200 and ScanNetV2. Notably, {\methodname} surpasses prior methods by \textbf{+8.6} and \textbf{+6.8} mAP on the validation and hidden test sets of ScanNet200, respectively.
\end{itemize}

\section{Related Works}
\noindent\textbf{3D Instance Segmentation}.
3D instance segmentation aims to identify, segment, and semantically categorize each object instance within a 3D scene. 
Existing approaches generally fall into three categories: proposal-based, grouping-based, and Transformer-based methods. 
Proposal-based methods~\cite{yang2019learning, hou20193d, yi2019gspn, engelmann20203d, td3d} are similar to their successful counterparts in 2D object detection, where the core idea is first to detect 3D bounding boxes and then refine them. 
In contrast, grouping-based methods~\cite{liang2021instance, chen2021hierarchical, vu2022softgroup, jiang2020pointgroup, wang2019associatively, jiang2020end, zhang2021point} do not explicitly generate object proposals. Instead, they directly cluster 3D points into distinct instances based on certain similarity criteria.

Recently, Transformer-based methods~\cite{spformer, al20233d, competitorformer} have emerged and started to dominate this field. Similar to DETR-like architectures in 2D perception, they directly predict segmentation results and offer advantages in both accuracy and inference speed. Mask3D~\cite{Mask3D} first proposes a Transformer architecture for 3D instance segmentation. QueryFormer~\cite{queryformer} and MAFT~\cite{maft} further improved query initialization and refinement. 
OneFormer3D~\cite{oneformer3d} unifies semantic, instance, and panoptic segmentation within a single framework. 
Relation3D~\cite{lu2025relation3d} explicitly models relationships through a geometric-biased self-attention for queries and a contrastive learning scheme for scene features.
These methods operate directly on 3D point clouds, while ignoring the use of 2D images, whereas ODIN~\cite{ODIN} leverages RGB-D images and employs a unified model to perform both 2D and 3D segmentation.

\noindent\textbf{Leverage 2D Models for 3D Segmentation}.
Recent open-vocabulary approaches attempt to incorporate 2D models into 3D segmentation. However, some methods~\cite{SAM3D, SAI3D} simply merge the mask predictions from 2D models such as SAM~\cite{SAM}, without leveraging intermediate 2D features. Others~\cite{DITR, openmask3d, openscene} utilize image features provided by CLIP~\cite{clip} or DINOv2~\cite{DINOv2} from independent views, but lack global 3D context fusion across these features.

\begin{figure*}[t]
    \centering
        \includegraphics[width=0.98\linewidth]{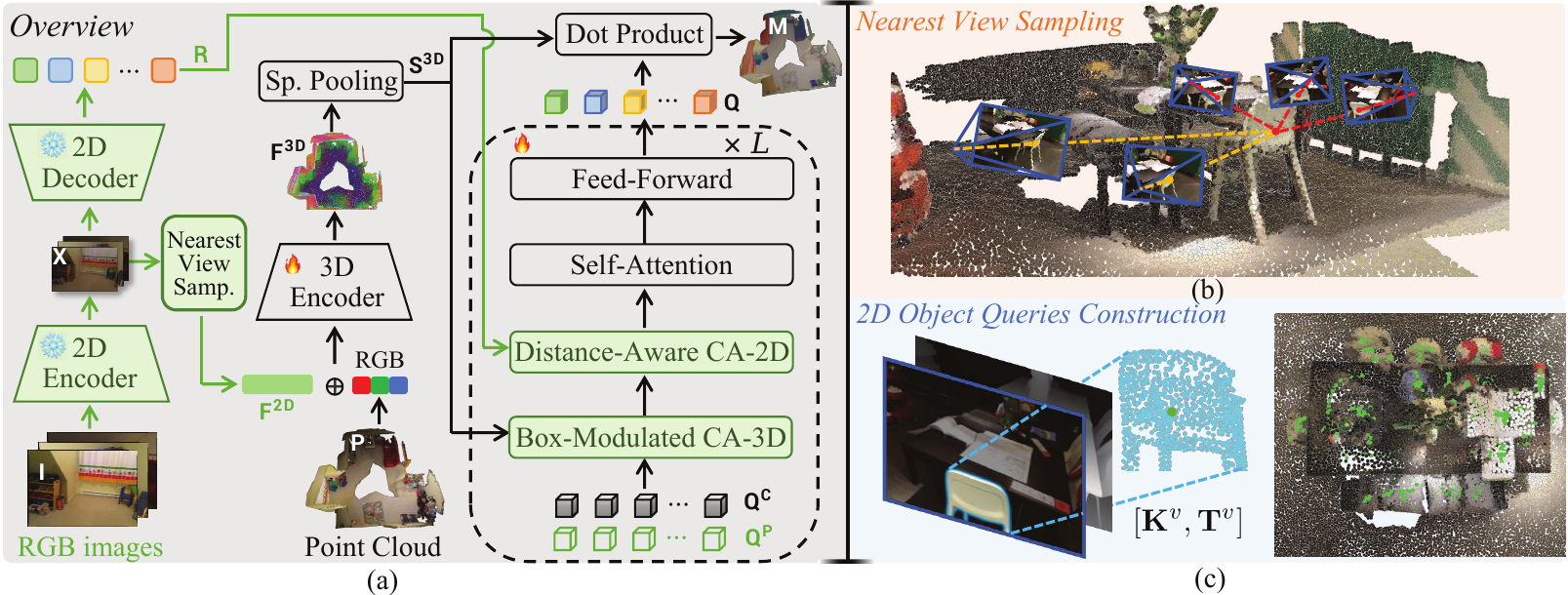}
    \caption{(a) Overview of {\methodname}, our contributions are highlighted in green. 
    (b) Visual illustration of the Nearest View Sampling operation. Each 3D point calculates its distance to all the views that it is visible to finds the top-k nearest views (red dash lines).  
    (c) Visual illustration of 2D object queries construction. 
     Each 2D object query is assigned a 3D center computed as the medoid of its corresponding 3D points, which are derived by projecting the depth values within its 2D mask using the camera parameters.
    The green points on the right side show the distribution of the 2D object queries' 3D centers in the scene.
    } 
    \label{fig.overview}
\end{figure*}

\section{Method}
\label{sec:method}

\subsection{Overview}
\label{sec:overview}
As shown in Fig.~\ref{fig.overview} (a), in addition to taking the point cloud $\mathbf{P} \in \mathbb{R}^{N\times 3}$ with $N$ points as input, {\methodname} also considers the point cloud's corresponding $V$ posed RGB images $\mathbf{I} \in \mathbb{R}^{V\times H\times W \times 3}$ and depth maps $\mathbf{D} \in \mathbb{R}^{V\times H\times W}$, where $H\times W$ denotes the image resolution.

In the encoder stage, we use a frozen 2D DETR-based model to process the images $\mathbf{I}$ and obtain the corresponding 2D image feature maps $\mathbf{X} \in \mathbb{R}^{V \times h \times w \times C}$, where $h \times w$ is the resolution of the feature maps\footnote{For clarity, we assume the feature maps are of single scale.} and $C$ is the number of channels. 
Then, as shown in Fig.~\ref{fig.overview} (b), we employ Nearest view Sampling to extract 2D features $\mathbf{F}^{2D} \in \mathbb{R}^{N\times C}$ from $\mathbf{X}$ for each 3D point. 
After fusing $\mathbf{F}^{2D}$ with the point cloud $\mathbf{P}$, we adopt a 3D encoder to perform global 3D context fusion, resulting in 3D point features $\mathbf{F}^{3D} \in \mathbb{R}^{N\times C}$ which encapsulate both rich 2D semantic information and detailed 3D geometric structure.
Finally, following the previous work~\cite{spformer}, we aggregate the 3D point features $\mathbf{F}^{3D}$ into superpoint features $\mathbf{S}^{3D} \in \mathbb{R}^{S \times C}$ using average pooling based on the pre-computed superpoints~\cite{landrieu2018large}.
Considering that the insufficiently trained 3D encoder may contaminate the high-quality 2D features, we further leverage the 2D object queries output by the 2D model.
We project the 2D object queries into the 3D space and perform downsampling to provide $O$ 2D object-level features, denoted as $\mathbf{R} \in \mathbb{R}^{O \times C}$, as illustrated in Fig.~\ref{fig.overview} (c).

In the decoder stage, we send the 3D superpoint features and the 2D object-level features to a Transformer decoder with $L$ layers for 3D instance segmentation.
Similar to the previous work~\cite{oneformer3d}, we randomly select $M$ superpoints as the initial 3D object queries during training, and initialize their content part with the corresponding superpoint features $\mathbf{S}^{3D}$, denoted as $\mathbf{Q}^{C} \in \mathbb{R}^{M \times C}$. 
While during evaluation, we regard all superpoints as initial 3D object queries, resulting in $M \equiv S$.
Meanwhile, different from prior works, we additionally utilize the positions of these $M$ superpoints to initialize their positional part $\mathbf{Q}^{P} \in \mathbb{R}^{M \times 6}$, where each represents an axis-aligned 3D bounding box in the scene. 
The content and positional components together constitute each 3D object query. 

In each decoder layer, the 3D object queries first perform Box-Modulated Cross-Attention (BMCA-3D) with the 3D superpoint features $\mathbf{S}^{3D}$. 
Then, based on the Distance-Aware attention mask, they cross-attend to the 2D object queries $\mathbf{R}$, which preserve the original 2D information, enabling direct injection of object-level 2D cues (DACA-2D).
Finally, the 3D object queries are refined by a self-attention layer followed by a feed-forward network to further enhance their feature representations.
After each decoder layer, the positional part of the 3D object queries is updated by linear regression.
After each decoder layer, the positional part of the 3D object queries is updated by linear regression.
And the 3D segmentation results $\mathbf{M}\in \{0, 1\}^{M\times S}$ are updated by computing the similarity between the updated content feature and the 3D superpoint features $\mathbf{S}^{3D}$.

\subsection{Feature Preparation in the Encoder Stage}
\label{sec:prepare}

Before introducing the novel modules in the decoder, we first describe how to construct the required 3D superpoint features $\mathbf{S}^{3D}$ and 2D object queries $\mathbf{R}$ in the encoder stage.

\subsubsection{3D Superpoint Feature Construction.}
We first inject semantic information from 2D images into the 3D point cloud through Nearest View Sampling. 
Specifically, given known camera parameters, we can project each 3D point onto multiple views and sample features at the projected location to provide a 2D feature for each 3D point.
However, a 3D point may be visible in hundreds of views, selecting a subset is crucial for computational efficiency.
Considering that the closer a 3D point is to the camera's optical center, the more details will be captured, we propose to find the top-$k$ (in our implementation, $k=3$) nearest views for each 3D point.
Without loss of generality, given the $i$-th 3D point's world coordinate as $[X_{i}^w, Y_{i}^w, Z_{i}^w]^\top$, its distance to $V$ views' camera centers $\bm{\rho}_i = [\rho_i^0, \rho_i^1, \dots, \rho_i^{V-1}]$ can be computed as
\begin{equation}
    \begin{gathered}
    \rho_i^v = \lVert \mathbf{T}^v [X_{i}^w, Y_{i}^w, Z_{i}^w, 1]^\top\lVert_2,
    \end{gathered}
\end{equation}
where $\mathbf{T}^v \in \mathbb{R}^{4 \times 4}$ is the extrinsic parameter of the $v$-th camera view, $\rho_i^v\in\mathbb{R}$ is the distance between the $i$-th 3D point and the $v$-th camera. The $i$-th 3D point's visibility $\bm{\delta}_i = [\delta_i^0, \delta_i^1, \dots, \delta_i^{V-1}]$ in $V$ views can be obtained by
\begin{equation}
    \begin{gathered}
        \relax [u_i^v, v_i^v, d_i^v] = \mathbf{K}^v\mathbf{T}^v [X^w_{i}, Y^w_{i}, Z^w_{i}, 1]^\top,\\
        \tilde{d}_i^v = \operatorname{Bili}(\mathbf{D}_v, [u_i^v, v_i^v]),\\
        \delta_i^v = (u_i^v \!\in\! (0,W)) \wedge (v_i^v \!\in\! (0,H)) \wedge (d_i^v \leq \tilde{d}_i^v),
    \end{gathered}
\end{equation}
where $\mathbf{K}^v\in\mathbb{R}^{3\times4}$ is the intrinsic parameter of the $v$-th camera view, augmented with a zero column for compatibility with homogeneous coordinates, $[u_i^v, v_i^v, d_i^v]$ is the coordinate of the $i$-th point in the $v$-th view's image coordinate system, $\tilde{d}_i^v\in\mathbb{R}$ is the depth value sampled on the $v$-th view's depth map, $\delta_i^v\in \{0, 1\}$ is the visibility of the $i$-th 3D point in the $v$-th view, and $\operatorname{Bili}()$ is the bilinear interpolation. Based on the distances and the visibilities, we obtain the indices of the top-$k$ nearest views $\operatorname{Nst}(\bm{\rho}_i, \bm{\delta}_i, k)$ for the $i$-th 3D point, where $\operatorname{Nst}()$ is the top-$k$ nearest selection function.

After identifying the $k$ nearest views for each 3D point, we sample and aggregate the corresponding 2D features by
\begin{equation}
    \begin{gathered}
    \mathbf{F}^{2D} = [\mathbf{f}^{2D}_{0}, \mathbf{f}^{2D}_{1}, \cdots, \mathbf{f}^{2D}_{N-1}],\\
    \mathbf{f}^{2D}_{i}= \frac{1}{k}\sum\nolimits_{v\in\operatorname{Nst}(\bm{\rho}_i, \bm{\delta}_i, k)} \operatorname{Bili}\left(\mathbf{X}_v,[u^v_i,v^v_i]\right),
    \end{gathered}
\end{equation}
where $\mathbf{f}_i^{2D} \in \mathbb{R}^C$ is the $i$-th 3D point's 2D feature, and $\mathbf{X}_v \in \mathbb{R}^{h \times w \times C}$ is the $v$-th view's 2D feature map. 

Subsequently, we enhance the 3D point cloud $\mathbf{P}$ with rich semantic cues from 2D by concatenating $\mathbf{F}^{2D}$ along the feature dimension.
However, since the 2D features $\mathbf{F}^{2D}$ are extracted from different and independent 2D images, they only encode information visible from specific viewpoints, lacking a global understanding of the scene and the spatial relationships among its elements.
To address this, we employ a 3D encoder to establish global contextual information for $\mathbf{F}^{2D}$, resulting in enhanced 3D features $\mathbf{F}^{3D}$.
Finally, we group $\mathbf{F}^{3D}$ into superpoint features $\mathbf{S}^{3D}$ according to pre-computed superpoints by average pooling.

\subsubsection{2D Object Query Construction.}
Since each scene corresponds to a large number of 2D images, it is impractical to take all 2D objects from every 2D image into consideration. To address this, we select only a subset.
We first only retain high-confidence 2D objects, and then project them into 3D space based on the depth values within their 2D masks and the corresponding camera parameters, resulting in a 3D point cloud for each 2D object.
To alleviate the effect of the depth map's noise, we adopt the Partitioning Around Medoids (PAM) algorithm~\cite{kaufman2009finding} to determine the 3D center of each 2D object.
Then, we apply the Farthest Point Sampling algorithm to downsample the 2D objects, ensuring that they are evenly distributed in 3D space and cover as many distinct instances as possible. This results in $O$ objects’ content features $\mathbf{R} \in \mathbb{R}^{O \times C}$ and their corresponding 3D centers $\mathbf{L}^o \in \mathbb{R}^{O \times 3}$.

It is worth highlighting that, instead of directly retrieving information from the original image feature maps via cross-attention, our constructed 2D object queries $\mathbf{R} \in \mathbb{R}^{O \times C}$ provide a significantly more memory-efficient implementation. 
In our implementation, for each 3D scene, we retain only $O = 2048$ queries, while the scene typically contains thousands of images. 
Although using the coarsest feature maps may reduce memory at the cost of information loss, the computational overhead remains high. 
By contrast, the queries provided by 2D models serve as both compact and semantic-rich object-level representations.

\subsection{Box-Modulated Cross-Attention with 3D Superpoint}
\label{sec:bmca-3d}
We propose box-modulated cross-attention, which modulates the attention map based on estimated object size. 
Specifically, we introduce a positional part $\mathbf{Q}^{P} \in \mathbb{R}^{M \times 6}$ for 3D object queries, each of them explicitly encodes the center coordinate and dimension of an axis-aligned 3D bounding box for an 3D instance, denoted as $(x, y, z, l, w, h)$.
We first introduce how conventional mask cross-attention with 3D superpoints is performed. 
Then, we describe the implementation of our proposed box-modulated cross-attention.

\subsubsection{Mask Cross-Attention with 3D Superpoint.}
Following the previous method~\cite{oneformer3d}, we employ mask attention here. 
We first calculate the segmentation results $\mathbf{M}\in \{0, 1\}^{M\times S}$ for all 3D object queries, and then transform the segmentation results to the attention masks $\mathbf{M}_\infty \in \mathbb{R}^{M\times S}$ to constrain the attention allocation. 
The process of obtaining the attention mask can be formulated as
\begin{equation}
    \begin{gathered}
        \mathbf{M} = \left[\sigma(\mathbf{Q}^{C} {\mathbf{S}^{3D}}^\top) > \tau_{\text{sim}}\right], 
        \mathbf{M}_{\infty} = \left(\mathbf{M} - 1\right) \cdot (\infty),
    \end{gathered}
    \label{eq:mask}
\end{equation}
where $\sigma$ is the sigmoid function and $\tau_{sim}$ is a similarity threshold. 
If the similarity between the $i$-th content query and the $j$-th superpoint is larger than $\tau_{sim}$, the superpoint will be assigned to the $i$-th object.
After that, we share the attention mask when querying features from the superpoint features $\mathbf{S}^{3D}$ in cross attention, which can be formulated as
\begin{equation}
    \begin{gathered}
        SimCont = \mathbf{Q}^{C} {\mathbf{S}^{3D}}^\top / \sqrt{C}, \\
        \mathbf{A} = \operatorname{SoftMax}(SimCont + \mathbf{M}_{\infty}), \\
        \mathbf{Q}^{C} \Leftarrow \mathbf{Q}^{C} + \mathbf{A}\mathbf{S}^{3D}, \\ 
    \end{gathered}
    \label{eq:ca-3d}
\end{equation}
where $SimCont$ indicates the content similarity between query and key, $\mathbf{A}\in\mathbb{R}^{M\times S}$ is the attention weights.
However, the attention mask $\mathbf{M}_\infty$ solely relies on the similarity between the content embeddings of the 3D object queries and the superpoint features, without considering any spatial constraints. 
This may result in high similarity scores between different instances of the same category that are spatially far apart, potentially leading to confusion. 
Some methods~\cite{maft, lu2025relation3d} attempt to incorporate positional embeddings to provide spatial constraints. However, it is still isotropic and insufficient for capturing object scale variations, which are prevalent in 3D environments.
Therefore, we further propose a box-modulated cross-attention mechanism that breaks the isotropy of positional embeddings, enabling more precise attention control.

\subsubsection{Box-Modulated Cross-Attention.}
Since each 3D object query contains a positional component that explicitly represents the center coordinate and size of the corresponding instance, we leverage the center coordinate to add positional embedding and use the size information to modulate the positional attention map.
First, by computing the dot product between the positional embeddings of queries and keys, we introduce an additional positional part to the query-to-key similarity, resulting in a change of attention weights: 
\begin{equation}
    \begin{gathered}
        SimPos= \operatorname{PE}(\mathbf{Q}_{[xyz]}^{P})  \operatorname{PE}(\mathbf{L}^{sp})^\top / \sqrt{C}, \\
        \mathbf{A} = \operatorname{SoftMax}(SimCont + SimPos + \mathbf{M}_{\infty}), \\
    \end{gathered}
\end{equation}
where $\mathbf{Q}_{[xyz]}^{P} \in \mathbb{R}^{M \times 3}$ are the center coordinates $(x,y,z)$ of $\mathbf{Q}^{P}$, $\mathbf{L}^{sp}\in\mathbb{R}^{S\times 3}$ are the center locations of superpoints, $SimPos$ is the positional similarity, and $\operatorname{PE}()$ is the positional encoding function. 
In implementation, we use sinusoidal positional encoding with a temperature of $T = 20$. 

Furthermore, we employ the dimensions $(l, w, h)$ of $\mathbf{Q}^{P}$ to modulate the positional attention maps. 
We define the modulation function $\operatorname{Mod}(a, b)$ as follows
\begin{equation}
    \begin{gathered}
        \mathbf{Q}^{ref} = \sigma(\operatorname{MLP}(\mathbf{Q}^{C})), \\
        \operatorname{Mod}(x, l) = (\operatorname{PE}(\mathbf{Q}_{[x]}^{P})  \operatorname{PE}(\mathbf{L}_{[x]}^{sp})^\top) \odot (\mathbf{Q}_{[l]}^{ref} / \mathbf{Q}_{[l]}^{P}), \\
    \end{gathered}
\end{equation}
where $\mathbf{Q}^{ref} \in \mathbb{R}^{M \times 3}$ is the reference value for the dimension of the bounding box, which is predicted by an MLP from $\mathbf{Q}^{C}$.
$\mathbf{Q}_{[x]}^{P} \in \mathbb{R}^{M \times 1}$ and $\mathbf{L}_{[x]}^{sp} \in \mathbb{R}^{S \times 1}$ are the $x$-coordinate of $\mathbf{Q}^{P}$ and $\mathbf{L}^{sp}$, respectively. 
$\odot$ denotes element-wise multiplication with broadcasting. 
$\mathbf{Q}_{[l]}^{ref} \in \mathbb{R}^{M \times 1}$ and $\mathbf{Q}_{[l]}^{P} \in \mathbb{R}^{M \times 1}$ are the $l$-dimension of $\mathbf{Q}_{l}^{ref}$ and $\mathbf{Q}_{[l]}^{P}$. The modulated positional similarity is computed as
\begin{equation}
    \begin{gathered}
        SimPos = \frac{\operatorname{Mod}(x, l) + \operatorname{Mod}(y, w) + \operatorname{Mod}(z, h)}{\sqrt{C}}. \\
    \end{gathered}
\end{equation}

This modulation allows the attention map to adapt based on the size of the objects, enhancing the model's ability to focus on relevant spatial regions.

\subsection{Distance-Aware Cross-Attention with 2D Object Queries}
\label{sec:daca-2d}
Given that a well-trained 2D DETR-like model is capable of providing semantic-rich 2D object queries, we propose to conduct cross-attention between 2D and 3D object queries.

\subsubsection{Distance-Aware Attention Mask Construction.}
Cross-attending to all 2D object queries from every 3D object query incurs extra computational cost and may cause confusion. Therefore, we introduce a distance-aware attention mask based on spatial proximity, allowing each 3D object query to only attend to nearby 2D object queries, which are more likely to belong to the same 3D instance.
However, since objects captured in 2D images are often partial observations, the 3D centers of the 2D objects may deviate significantly from their corresponding 3D objects' centers. 
As a result, directly computing the attention mask based on the 3D spatial distance between 3D object queries and 2D objects may bring noise. We propose to first find the spatial relationship between 2D objects and superpoints, then, according to the assignment relationship between superpoints and 3D object queries (from segmentation results $\mathbf{M}$), we obtain the attention mask between 3D object queries and 2D objects $\mathbf{M}^{o}_\infty \in \mathbb{R}^{M\times O}$. The process can be formulated as
\begin{equation}
\begin{gathered}
    dist = \operatorname{Dist}(\mathbf{L}^o, \mathbf{L}^{sp}), ~~ \mathbf{M}^d = [dist < \tau_{dist}], \\
    \mathbf{M}^o = \mathbf{M} \mathbf{M}^d, ~~
    \mathbf{M}_\infty^o = \left(\mathbf{M}^o - 1\right)\cdot(\infty),
\end{gathered}
\end{equation}
where $\operatorname{Dist}()$ is an operator that calculates the $L2$ distance $dist\in \mathbb{R}^{S\times O}$ between every 3D superpoint and 2D object, $\tau_{dist}$ is the distance threshold to determine whether a 3D superpoint and a 2D object are related.

\subsubsection{Cross-Attention with 2D Object Queries.}
Based on the attention mask, we further optimize the 3D content queries with the 2D object-level features through cross-attention
\begin{equation}
    \begin{gathered}
        \mathbf{A}^{o} = \operatorname{SoftMax}(\mathbf{Q}^{C}\mathbf{R}^\top / \sqrt{C} + \mathbf{M}^o_{\infty}), \\
        \mathbf{Q}^{C} \Leftarrow \mathbf{Q}^{C} + \mathbf{A}^{o}\mathbf{R}, \\ 
    \end{gathered}
\end{equation}
where $\mathbf{A}^o \in \mathbb{R}^{M\times O}$ is the attention weight.

\setlength{\tabcolsep}{3.2pt}
\begin{table*}[t]
    \centering
    \begin{tabular}{l|ccc|ccc|ccc|ccc}
    \shline
     & \multicolumn{3}{c|}{All} & \multicolumn{3}{c|}{Head} & \multicolumn{3}{c|}{Common} & \multicolumn{3}{c}{Tail} \\
    Method & mAP & mAP$_{50}$ & mAP$_{25}$ & mAP & mAP$_{50}$ & mAP$_{25}$ &  mAP & mAP$_{50}$ & mAP$_{25}$ &  mAP & mAP$_{50}$ & mAP$_{25}$ \\
    \shline
    \rowcolor{gray!25}
    \multicolumn{13}{l}{\textbf{\textit{Validation Set}}} \\
    OpenMask3D~\shortcite{openmask3d} & 15.4 & 19.9 & 23.1 & - & - & - & - & - & - & - & - & -\\
    Mask3D~\shortcite{Mask3D} & 27.4 & 37.0 & 42.3 & 40.3 & 55.0 & 62.2 & 22.4 & 30.6 & 35.4 & 18.2 & 23.2 & 27.0\\
    MAFT~\shortcite{maft} & 29.2 & 38.2
    & 43.3 & - & - & - & - & - & - & - & - & -\\
    SAI3D~\shortcite{SAI3D} & 12.7 & 18.8 & 24.1 & 12.1 & - & - & 10.4 & - & - & 16.2 & - & -\\
    Open3DIS~\shortcite{open3dis} & 23.7 & 29.4 & 32.8 & 27.8 & - & - & 21.3 & - & - & 21.8 & - & -\\
    Open-YOLO 3D~\shortcite{openyolo3d} & 24.7 & 31.7 & 36.2 & 27.8 & - & - & 24.3 & - & - & 21.6 & - & -\\
    OneFormer3D~\shortcite{oneformer3d} & 30.2 & 40.9 & 44.6 & 42.0 & 57.7 & 63.9 & 27.0 & 36.3 & 39.8 & 20.1 & 26.6 & 27.7 \\
    ODIN~\shortcite{ODIN} & 31.5 & 45.3 & 53.1 & 37.5 & 54.2 & 66.1 & 31.6 & 43.9 & 50.2 & 24.1 & 36.6 & 41.2\\
    Relation3D~\shortcite{lu2025relation3d}& 31.6 & 41.2 & 45.6 & - & - & - & - & - & - & - & - & -\\
    \toprule
    Ours-Grounding DINO & 36.2 & 47.6 & 54.2 & 44.8 & 60.9 & 69.6  & 32.0 & 42.3 & 47.3 & 31.1 & 38.2 & 44.4 \\
    Ours-DINO-X& \textbf{40.2} & \textbf{52.4} & \textbf{58.6} & \textbf{46.3} & \textbf{63.7} & \textbf{71.5} & \textbf{37.4} & \textbf{47.6} & \textbf{52.3} & \textbf{36.2} & \textbf{44.9} & \textbf{51.0} \\
    \shline
    \rowcolor{gray!25}
    \multicolumn{13}{l}{\textbf{\textit{Hidden Test Set}}} \\
    Mask3D~\shortcite{Mask3D} & 27.8 & 38.8 & 44.5 & 38.3 & 54.2 & 65.3 & 26.3 & 35.7 & 39.2 & 16.8 & 23.7 & 25.4 \\
    TD3D~\shortcite{td3d}   & 21.1 & 32.0 & 37.9 & 33.2 & 50.1 & 60.3 & 17.7 & 26.4 & 30.6 & 10.3 & 16.4 & 19.0 \\
    ODIN~\shortcite{ODIN}   & 26.5 & 38.1 & 45.1 & 34.9 & 50.7 & 63.7 & 26.8 & 37.5 & 40.7 & 16.3 & 23.7 & 27.7\\
    \toprule
    Ours-DINO-X & \textbf{34.6} & \textbf{45.4} & \textbf{51.1} & \textbf{43.7} & \textbf{58.7} & \textbf{68.5} & \textbf{35.3} & \textbf{45.3} & \textbf{48.4} & \textbf{22.9} & \textbf{29.6} & \textbf{33.1} \\
    \shline
    \end{tabular}
    \caption{Comparison of {\methodname} with prior methods on validation set and hidden test set of ScanNet200.} 
    \label{tab:benchmark_200_full}
\end{table*}
\setlength{\tabcolsep}{6pt}

\subsection{Updating of Mask and Box}
\label{sec:update}
The mask update follows Equation~\ref{eq:mask} in Sec.~\ref{sec:bmca-3d}. 
For each box query, the center coordinate $(x, y, z)$ is initialized using the coordinates of its corresponding superpoint, while the dimension $(l, w, h)$ is initialized with fixed values.
After each decoder layer, we apply an MLP to predict a residual offset, enabling layer-by-layer refinement of $\mathbf{Q}^{P}$.
The update process at the $l$-th decoder layer can be formulated as
\begin{equation}
    \begin{gathered}
        \mathbf{Q}^{P}_{l+1} = \mathbf{Q}^{P}_{l} + \operatorname{MLP}(\mathbf{Q}^{C}_{l+1}).\\
    \end{gathered}
\end{equation}

To supervise the 3D positional queries $\mathbf{Q}^{P}$, we set up an additional regression task. 
We construct axis-aligned box labels for all the instances based on the ground truth masks, and the predicted boxes are supervised by an L1 loss. 
Moreover, we incorporate a bounding box regression cost into the matching strategy during training.

Finally, it is worth noting that with negligible computational overhead, we can use the predicted box of each 3D object query to filter out mask regions that are far from the box, thereby further improving segmentation accuracy.

\section{Experiments}
\label{sec:exp}

\subsection{Experiment Setup}
\label{sec:setup}

\noindent\textbf{Dataset.}
We conduct experiments on the ScanNetv2~\cite{scannet} and ScanNet200~\cite{scannet200} datasets. 
ScanNetv2 is a richly annotated 3D indoor scene reconstruction dataset, containing 20 semantic classes and annotations for 18 object instances. 
It provides point clouds, RGB-D images, and corresponding camera intrinsics and extrinsics. 
The dataset is divided into training, validation, and hidden test splits, consisting of 1201, 312, and 100 scenes, respectively.
ScanNet200 extends the original ScanNetv2 by significantly increasing the number of instance categories, providing 198 instance-level classes and 2 semantic classes. 
According to the number of annotations, these 200 classes are further categorized into ``head'', ``common'', and ``tail'' splits, with 66, 68, and 66 categories respectively. 
This long-tailed distribution, with the head classes being more frequent and the tail classes more sparse, makes ScanNet200 a more challenging benchmark, especially for evaluating model performance on real-world long-tail distributions. 
To validate the superiority of {\methodname}, we report the detailed performance on the 3 sub-splits of ScanNet200.

\noindent\textbf{Evaluation Metrics}.
We adopt the standard evaluation metric for 3D instance segmentation, mean Average Precision (mAP), which measures the area under the precision-recall curve across different Intersection-over-Union (IoU) thresholds. 
Specifically, mAP is computed by averaging the precision values at IoU thresholds ranging from 50\% to 95\%, with an interval of 5\%, providing a comprehensive assessment of segmentation quality. 
In addition to the overall mAP, we also report mAP$_{50}$ and mAP$_{25}$, which reflect model performance at fixed IoU thresholds of 50\% and 25\%, respectively. 
These metrics are particularly useful for analyzing performance under different levels of localization tolerance.

\noindent\textbf{Implementation Details}.
For the hyperparameter settings, we primarily follow OneFormer3D~\cite{oneformer3d} and adopt it as our baseline.
Specifically, we voxelize the 3D points with a voxel size of 2cm and employ a U-Net-like sparse convolutional network as our 3D encoder~\cite{minkowski} to globally encode the point cloud enhanced by image-level features.
During training, we use the AdamW~\cite{loshchilov2017decoupled} optimizer with an initial learning rate of $1 \times 10^{-4}$, a weight decay of 0.05, a batch size of 4, and a polynomial learning rate scheduler with a base of 0.9. 
We adopt the disentangled matching strategy proposed in OneFormer3D, and retain the cross-entropy loss $\mathcal{L}_{cls}$ for classification, the binary cross-entropy loss $\mathcal{L}_{bce}$ and Dice loss $\mathcal{L}_{dice}$ for the superpoint mask, as well as the binary cross-entropy loss $\mathcal{L}_{sem}$ for semantic. 
We also employ the same data augmentation strategies, including horizontal flipping, random rotations around the z-axis, elastic distortion, and random scaling.

Distinct from previous methods, {\methodname} achieves significantly faster convergence, reducing the number of training epochs from 432 to 88, which results in about a $5\times$ reduction in training time (see Fig.~\ref{fig.teaser}(b)). 
In addition, we develop two versions, incorporating either Grounding DINO~\cite{groundingdino} or DINO-X~\cite{DINOX} as the 2D model to provide 2D features and queries. 
For the box regression task, we further introduce an additional L1 loss $\mathcal{L}_{box}$. Therefore, the overall loss function is:
\begin{equation}
    \mathcal{L} = \lambda_{1} \cdot \mathcal{L}_{cls} + \mathcal{L}_{bce} + \mathcal{L}_{dice} + \lambda_{2} \cdot \mathcal{L}_{sem} + \lambda_{3} \cdot \mathcal{L}_{box}
\end{equation}
where $\lambda_{1} = \lambda_{2} = \lambda_{3} = 0.5$ in our implementation. 
Both training and inference of {\methodname} can be performed on a single NVIDIA RTX 3090 GPU.

\subsection{Comparison with Prior Work}
\label{benchmark}

We compare our method with previous approaches on the ScanNet200 validation and hidden test set. 
Our method offers two versions that utilize either Grounding DINO or DINO-X as the 2D model to provide features. In the following discussions, we refer to the DINO-X-based version for comparison.
As shown in Table~\ref{tab:benchmark_200_full}, on the validation set, {\methodname} outperforms all existing methods by a large margin in all sub-splits, achieving a +8.6 mAP higher mAP than the previous best method. 
Especially, {\methodname}'s performance on the tail-split even surpasses previous methods on the common-split (36.2 mAP vs. 31.6 mAP). 
This indicates that the incorporation of 2D features significantly alleviates the issue of insufficient 3D training data for long-tailed categories.
Meanwhile, on the hidden test set, {\methodname} also shows significant improvement, establishing a new state of the art.
These consistent and substantial improvements demonstrate the superiority of {\methodname}.

\setlength{\tabcolsep}{6pt}
\begin{table}[h]
\centering
\begin{tabular}{c|ccc}
\toprule
Method  & mAP & mAP$_{50}$ & mAP$_{25}$  \\
\midrule
SSTNet~\shortcite{liang2021instance} &      49.4    &     64.3      & 74.0  \\
SoftGroup~\shortcite{vu2022softgroup} &      46.0    &     67.6      & 78.9  \\
Mask3D~\shortcite{Mask3D} &      55.2        &     73.7      & 83.5  \\
SPFormer~\shortcite{spformer} &  56.3        &     73.9      & 82.9  \\
QueryFormer~\shortcite{queryformer} &      56.5    &     74.2      & 83.3  \\
MAFT~\shortcite{maft} &      59.9    &     76.5      & -  \\
OneFormer3D~\shortcite{oneformer3d} &      59.3    &     78.1      & 86.4  \\
TD3D~\shortcite{td3d} &      47.3    &     71.2      & 81.9 \\
ODIN~\shortcite{ODIN} &      50.0    &     71.0      & 83.6  \\
Spherical Mask~\shortcite{shin2024spherical} &      62.3    &     79.9      & 88.2  \\
Relation3D~\shortcite{lu2025relation3d} &      62.5    &     80.2      & 87.0  \\
Ours-DINO-X & \textbf{64.0}     & \textbf{81.5}     & \textbf{88.9} \\
\bottomrule
\end{tabular}
\caption{Comparison of {\methodname} with prior methods on the validation set of ScanNetv2.}
\label{tab:benchmark_20}
\end{table}
\setlength{\tabcolsep}{6pt}

Moreover, we compare {\methodname} with prior methods on the ScanNetv2 validation set.
As shown in Table~\ref{tab:benchmark_20}, our method outperforms all existing methods across all metrics. Specifically, it outperforms our baseline by +4.7 mAP and surpasses the previous state-of-the-art method by +1.5 mAP.
Compared with the improvement achieved on the more challenging ScanNet200, the improvement on ScanNetv2 is relatively smaller. 
We attribute this to the fact that 2D representations provide more discriminative semantic features, which are particularly advantageous under the fine-grained 200-class setting of ScanNet200.

\setlength{\tabcolsep}{4pt}
\begin{table}[t]
    \centering
    \begin{tabular}{l|c|c|c|ccc}
    \toprule
    Row & Img. F. & Obj. F. & Mod. & mAP & mAP$_{50}$ & mAP$_{25}$ \\
    \midrule
    1 & \ding{55}   & \ding{55}   & \ding{55}  & 26.8 & 36.9 & 42.1 \\
    2 & \checkmark   & \ding{55} & \ding{55}  & 35.2 & 46.9 & 54.6 \\
    3 & \ding{55}  & \checkmark   & \ding{55}  & 33.0 & 46.6 & 54.9 \\  
    4 & \checkmark  & \checkmark  & \ding{55}  & 38.3 & 50.5 & 57.4 \\ 
    5 & \checkmark  & \checkmark  & \checkmark  & 39.8 & 52.1 & 58.6 \\
    6$^\dag$ & \checkmark  & \checkmark  & \checkmark  & \textbf{40.2} & \textbf{52.4} & \textbf{58.6} \\ 
    \bottomrule
    \end{tabular}
    \caption{Ablation on our key components. 
    Img. F. denotes the use of 2D image-level features in the encoder,
    Obj. F. refers to the incorporation of 2D object-level features in the decoder, 
    and Mod. indicates the use of box modulated cross-attention. 
    $^\dag$ denotes using predicted boxes to filter masks.
    } 
    \label{tab:ablation_main}
\end{table}
\setlength{\tabcolsep}{6pt}

\setlength{\tabcolsep}{6pt}
\begin{table}[t]
\centering
\begin{tabular}{c|ccc}
\toprule
Methods  & mAP & mAP$_{50}$ & mAP$_{25}$  \\
\midrule
w/o Global Context Fusion &       35.9        &     48.7      & 55.4  \\
w Global Context Fusion & \textbf{39.8}     & \textbf{52.1}     & \textbf{58.6} \\
\bottomrule
\end{tabular}
\caption{Ablation on Global Context Fusion.}
\label{tab:ablation_earlyfuse}
\end{table}
\setlength{\tabcolsep}{6pt}

\setlength{\tabcolsep}{6pt}
\begin{table}[H]
\centering
\begin{tabular}{c|ccc}
\toprule
Sampling Methods  & mAP & mAP$_{50}$ & mAP$_{25}$  \\
\midrule
Farthest Sampling  &       36.2        &     48.5      & 56.3  \\
Nearest Sampling & \textbf{39.8}     & \textbf{52.1}     & \textbf{58.6} \\
\bottomrule
\end{tabular}
\caption{Ablation on Nearest View Sampling.}
\label{tab:ablation_sample}
\end{table}
\setlength{\tabcolsep}{6pt}

\subsection{Ablation Studies and Analysis}
\label{ablation}

We conduct ablations on the challenging ScanNet200 validation set to assess the effectiveness of our key components.

\noindent\textbf{Key Components.} 
We progressively add components to the baseline to validate their effectiveness. 
As shown in rows 1$\sim$3 of Table~\ref{tab:ablation_main}, incorporating globally enhanced 2D image-level features and leveraging information directly from raw 2D object-level features lead to improvements of +8.4 and +6.2 mAP, respectively. 
Moreover, thanks to their complementarity, where the image-level features enhance the 3D point cloud scene with 2D semantic information, while the object-level features directly inject the rich information from the original 2D object queries into the 3D object queries, combining both brings an additional gain of +3.1 mAP, as shown in rows 2 and 4.
Subsequently, as illustrated in rows 4 and 5, introducing box-modulated cross-attention using the estimated boxes yields a further gain of +1.5 mAP. 
Finally, as shown in rows 5 and 6, using the estimated boxes to filter out distant masks also brings a slight improvement.

\noindent\textbf{Global Context Fusion.} 
After establishing the correspondence between each 3D point and 2D image-level features, we apply a 3D encoder for global context fusion.
Alternatively, if we simply fuse the corresponding 2D image-level features to the point cloud through an MLP, as reported in Table~\ref{tab:ablation_earlyfuse}, there will be a -3.9 mAP performance drop, highlighting the effectiveness of global contextualization.

\noindent\textbf{Nearest View Sampling.} 
As discussed in Sec.\ref{sec:prepare}, we compare two sampling strategies: selecting the nearest $k$ views and selecting the farthest $k$ views. 
As shown in Table~\ref{tab:ablation_sample}, the ``nearest sampling'' strategy yields superior performance, suggesting that 2D information from views closer to the camera's optical center tends to be more informative.

\setlength{\tabcolsep}{5pt}
\begin{table}[!t]
\centering
\begin{tabular}{c|ccc}
\toprule
Methods  & mAP & mAP$_{50}$ & mAP$_{25}$  \\
\midrule
w/o Distance-Aware Mask  &       36.3        &     48.7      & 54.9  \\
w Distance-Aware Mask   & \textbf{39.8}     & \textbf{52.1}     & \textbf{58.6} \\
\bottomrule
\end{tabular}
\caption{Ablation on Distance-Aware Mask.}
\label{tab:ablation_queryattnmask}
\end{table}
\setlength{\tabcolsep}{6pt}

\noindent\textbf{Distance-Aware Attention Mask in DACA-2D.} 
As shown in Table~\ref{tab:ablation_queryattnmask}, when extracting information from 2D object queries, employing our proposed distance-aware mask significantly improves performance by selectively attending to the most relevant queries based on spatial relationships, leading to a +3.5 mAP improvement.

\setlength{\tabcolsep}{6pt}
\begin{table}[t]
    \centering
    \begin{tabular}{l|c|c|c|ccc}
    \toprule
    Row & Center PE. & Box Reg. & Box Mod. & mAP  \\
    \midrule
    1 & \ding{55}   & \ding{55}   & \ding{55}  & 38.3  \\
    2 & \checkmark   & \ding{55} & \ding{55}  & 38.1  \\
    3 & \checkmark  & \checkmark  & \ding{55}  & 37.8  \\ 
    4 & \checkmark  & \checkmark  & \checkmark  & \textbf{39.8} \\
    \bottomrule
    \end{tabular}
    \caption{Ablation on box queries. Center PE. indicates using positional embedding with center coordinates, Box Reg. indicates adding box regression loss, and Box Mod. indicates using box to modulate positional attention map.
    } 
    \label{tab:ablation_boxquery}
\end{table}
\setlength{\tabcolsep}{6pt}

\noindent\textbf{Box Modulation in BMCA-3D.} 
To validate the specific contributions of our proposed box-modulated cross-attention, we progressively integrate components in the order of implementation. 
As shown in Row 2 of Table~\ref{tab:ablation_boxquery}, we first introduce positional queries and add a center-point regression task, using the center coordinates to incorporate position embedding into the cross-attention. 
In row 3, we further add a box size regression task. 
It is observed that neither of these additions yields a significant performance improvement. 
Finally, we modulate the positional part of the cross-attention using the estimated box, resulting in a 1.5 mAP improvement, demonstrating its effectiveness.

\section{Conclusion}
In this paper, we have presented {\methodname}, a novel method for 3D instance segmentation.
{\methodname} comprehensively leverages both 2D image-level and object-level features from pre-trained 2D DETR-based detection models, effectively transferring the strong capabilities of 2D models into the 3D domain.
For the 2D image-level features, we select the most suitable views for each 3D point to extract features, aggregate them, and then perform global context fusion.
For the 2D object-level features, through a carefully designed distance-aware mask, we can effectively extract semantic-rich 2D information into 3D queries in a memory-efficient manner.
Finally, we introduce 3D box queries and modulate the positional attention maps using the estimated boxes to further enhance the precision of perception ability.
{\methodname} achieves state-of-the-art results on the 3D instance segmentation benchmarks ScanNetv2 and ScanNet200, and converges much faster during training, demonstrating its superiority.

\appendix

\bibliography{aaai2026}

\clearpage
\input{appendix}

\end{document}

%% file: appendix.tex
\setcounter{secnumdepth}{2}
\renewcommand{\thesection}{\Alph{section}}
\renewcommand{\thesubsection}{\thesection.\arabic{subsection}}

\section{Technical Appendix}

In the technical appendix, we first provide an efficiency analysis of {\methodname} (Sec.~\ref{sec_efficiency}).
Moreover, we conduct additional ablation studies based on the Grounding DINO to validate the effectiveness and generalization ability of our method (Sec.~\ref{sec_moreabl}). 
Furthermore, we provide more implementation details to facilitate reproducibility (Sec.~\ref{sec:more_details}).
Finally, we provide more qualitative (Sec.~\ref{sec_vis}) comparisons with previous methods to further demonstrate the superiority of our approach.

\subsection{Efficiency Analysis}
\label{sec_efficiency}
We evaluate the efficiency of our model on the ScanNet200 validation scenes and report the averaged results. 
We divide {\methodname} into two stages, namely Feature Preparation Stage and Model Inference Stage, and provide a detailed analysis of each.

\noindent\textbf{Feature Preparation Stage.} 
We first employ DINO-X~\cite{DINOX} to extract features from scene images. 
For each scene, we sample input frames at intervals of 5. 
When using RGB images at their original resolution with $width=1536$, the average processing time per scene is 94.3 s. 
By reducing the input resolution to $width=768$, the average time decreases to only 22.3 s, while performance remains nearly unchanged (a drop of merely -0.5 mAP). 
Subsequently, we apply the Nearest View Sampling strategy described in Sec.~\ref{sec:prepare} to extract appropriate features for each 3D point. 
This process involves simple matrix operations and takes only 0.8 ms per scene, which is negligible.

\noindent\textbf{Model Inference Stage.} 
Our model introduces only a marginal computational overhead compared to the baseline (OneFormer3D). 
Specifically, during inference on an RTX~3090 GPU, the per-scene processing time increases slightly from 310 ms to 350 ms, indicating that the proposed method maintains high efficiency while achieving superior performance. 
Similarly, the peak GPU memory reserved shows a minimal rise from 15.0 GB to 15.5 GB, demonstrating that the method remains memory-efficient. 
The detailed comparison is summarized in Table~\ref{tab:runtime_analysis}.

\setlength{\tabcolsep}{6pt}
\begin{table}[h!]
\centering
\begin{tabular}{lcc}
\toprule
Metric & Baseline & Ours \\
\midrule
Runtime per scene (ms) & 310 & 350 \\
Peak GPU memory reserved (GB)   & 15.0 & 15.5 \\
\bottomrule
\end{tabular}
\caption{Inference efficiency comparison with the baseline.}
\label{tab:runtime_analysis}
\end{table}
\setlength{\tabcolsep}{6pt}

In summary, in the offline scenarios, the runtime and memory consumption of {\methodname} are both practical.

\subsection{More Ablation Studies}
\label{sec_moreabl}
In this section, we present additional ablation studies to validate the effectiveness of our method. 
Following the main paper, all ablations are conducted on ScanNet200, and unless otherwise specified, we employ DINO-X~\cite{DINOX} as the 2D model.

\noindent\textbf{Utilization of Original Feature Maps.} 
In Sec.\ref{sec:intro}, we mention that integrating globally-enhanced 2D features in the encoder stage, while simultaneously preserving their semantic discrimination ability in the decoder stage, could yield significant improvements. We validate it in Sec.\ref{sec:exp}. 
Then, we also discuss different ways of leveraging the original 2D features in the decoder. 
Although directly attending to 2D image feature maps to update 3D queries (red lines in Fig.\ref{fig.teaser}(a)), as done in BEVFormer~\cite{li2022bevformer}, is generally more effective, it is not practical in the context of 3D scene perception, where each scene is generally reconstructed from thousands of images. 
Mapping 2D image features to discrete 3D points and updating 3D queries by attending to these discrete 2D features (green lines in Fig.\ref{fig.teaser}(a)) is practical, however, not effective.
Thus, we propose the Distance-Aware Cross-Attention with 2D Object Queries ({\queryca}) module. {\queryca} regards the 2D object queries from the decoder of 2D DETR-like models as a compact representation of 2D images and updates 3D queries by attending to the sparse 2D object queries, providing a memory-friendly way to attend to the original 2D features.

To validate the effectiveness, we conduct an ablation study. Considering practical feasibility, we only compare with the way that maps 2D image features to discrete 3D points. 
More specifically, we project the original 2D feature maps onto discrete 3D points and fuse them with the point cloud features $\mathbf{F}^{3D}$ obtained from the 3D encoder using an MLP. 
As shown in Table~\ref{tab:ablation_orimaps}, compared with the use of discrete 2D features, the use of the 2D object queries shows a significant superiority (+4.1 mAP). 

\setlength{\tabcolsep}{6pt}
\begin{table}[h]
\centering
\begin{tabular}{c|ccc}
\toprule
Methods  & mAP & mAP$_{50}$ & mAP$_{25}$  \\
\midrule
Mapping to 3D Points  &       35.7        &     47.2      & 54.3  \\
2D Object Queries   & \textbf{39.8}     & \textbf{52.1}     & \textbf{58.6} \\
\bottomrule
\end{tabular}
\caption{Ablation on Utilization of Original Feature Maps.}
\label{tab:ablation_orimaps}
\end{table}
\setlength{\tabcolsep}{6pt}

\noindent\textbf{Dropout of 2D Object-Level Queries.} 
As described in Sec.\ref{sec:more_details}, during training, we randomly drop $70\%$ of the 2D object queries as a form of data augmentation. As shown in Table \ref{tab:ablation_dropout}, we conduct an ablation study on this design. The results show that applying dropout slightly improves the model's performance, yielding a gain of +0.6 mAP.

\setlength{\tabcolsep}{6pt}
\begin{table}[ht]
\centering
\begin{tabular}{c|ccc}
\toprule
Methods  & mAP & mAP$_{50}$ & mAP$_{25}$  \\
\midrule
Without Dropout &       39.2        &     \textbf{52.1}      & 58.2  \\
With $70\%$ Dropout & \textbf{39.8}     & \textbf{52.1}     & \textbf{58.6} \\
\bottomrule
\end{tabular}
\caption{Ablation on Dropout of Object-Level Queries.}
\label{tab:ablation_dropout}
\end{table}
\setlength{\tabcolsep}{6pt}

\noindent\textbf{Key Components with Grounding DINO.} 
We use Grounding DINO~\cite{groundingdino} as the 2D DETR-based model to conduct the same ablation studies as in Table~\ref{tab:ablation_main}, to validate the effectiveness and generalization ability of our proposed key components. 
However, different from the setting in Sec.~\ref{ablation} of the main text, Grounding DINO is not as powerful as DINO-X (it is also trained on a smaller amount of 2D data), so we require a longer training schedule to achieve convergence. Specifically, we train for 108 epochs. 
As shown in Table~\ref{tab:ablation_main_gdino}, compared to the results in Table~\ref{tab:ablation_main}, the conclusions remain consistent.

\setlength{\tabcolsep}{4pt}
\begin{table}[htb]
    \centering
    \begin{tabular}{l|c|c|c|ccc}
    \toprule
    Row & Img. F. & Obj. F. & Mod. & mAP & mAP$_{50}$ & mAP$_{25}$ \\
    \midrule
    
    1 & \ding{55}   & \ding{55}   & \ding{55}  & 27.7 & 37.9 & 42.8 \\
    2 & \checkmark   & \ding{55} & \ding{55}  & 33.4 & 44.5 & 51.1 \\
    3 & \ding{55}  & \checkmark   & \ding{55}  & 31.1 & 42.7 & 49.1 \\  
    4 & \checkmark  & \checkmark  & \ding{55}  & 34.6 & 46.2 & 52.5 \\ 
    5 & \checkmark  & \checkmark  & \checkmark  & 36.0 & 47.5 & 54.2 \\
    6$^\dag$ & \checkmark  & \checkmark  & \checkmark  & \textbf{36.2} & \textbf{47.6} & \textbf{54.2} \\
    \bottomrule
    \end{tabular}
    \caption{Ablation on our key components with Grounding DINO as 2D model. 
    Img. F. denotes the use of 2D image-level features in the encoder,
    Obj. F. refers to the incorporation of 2D object-level features in the decoder, 
    and Mod. indicates the use of box modulated cross-attention. 
    $^\dag$ denotes using predicted boxes to filter masks.
    } 
    \label{tab:ablation_main_gdino}
\end{table}
\setlength{\tabcolsep}{6pt}

\subsection{More Implementation Details}
\label{sec:more_details}
In this section, we provide additional implementation details for our method introduced in Sec.~\ref{sec:method}.

\noindent\textbf{Utilization of 2D Image-Level Features.} 
As described in Sec.~\ref{sec:prepare}, for each 3D point, we sample image-level features from the top-$k$ nearest views and aggregate them to obtain the associated 2D image-level features $\mathbf{F}^{2D} \in \mathbb{R}^{N \times C}$. 
These features are then fused via a global context fusion module through a 3D encoder. 

In our implementation, we set $k=3$ for computational efficiency. 
Also, to improve training efficiency, we precompute $\mathbf{F}^{2D}$ for each scene in ScanNet200 and store them on disk for direct loading during training.

\noindent\textbf{Utilization of 2D Object-Level Queries.} 
For the 2D object-level queries $\mathbf{R} \in \mathbb{R}^{O \times C}$ constructed as presented in Sec.~\ref{sec:prepare}, we also precompute and store them on disk for loading during training. 
For each scene, we apply the Farthest Point Sampling (FPS) algorithm to downsample these 2D object queries to $O=2048$ based on their 3D locations. 
In edge cases, such as particularly small scenes where the total number of 2D object queries is fewer than 2048, we retain all available queries without downsampling. 

Moreover, to construct the 2D object queries, we need to project each 2D object with high confidence in the image into the 3D space. 
We explored various strategies to represent their 3D positions. 
Initially, we projected all 2D pixels belonging to the same 2D object into 3D space using camera parameters and a depth map. 
We then simply averaged the coordinates of these projected points to serve as the object's 3D center. 
However, we found that this approach was significantly susceptible to depth map noise, leading to inaccurate 3D center coordinates. 
Specifically, depth values at object boundaries tend to be blurry and can erroneously take on background depth values (suddenly becoming very large), causing the projected 3D points to become outliers. 
To address this issue, we adopted the Partitioning Around Medoids (PAM) algorithm~\cite{kaufman2009finding}, which is more robust to outliers, to estimate the 3D position of each 2D object. 
Specifically, we select the medoid, i.e., an actual projected 3D point that minimizes the total distance to all other projected points of the same object. 
This effectively reduces the impact of noisy depth values and yields more reliable object positions in 3D space.

Finally, it is important to note that during training, we randomly drop $70\%$ of the 2D object queries within each scene as a form of data augmentation to enhance model robustness. 
During inference, we retain all 2D object queries without dropout to leverage richer 2D object-level information.

\begin{figure*}[t]
    \centering
        \includegraphics[width=0.98\linewidth]{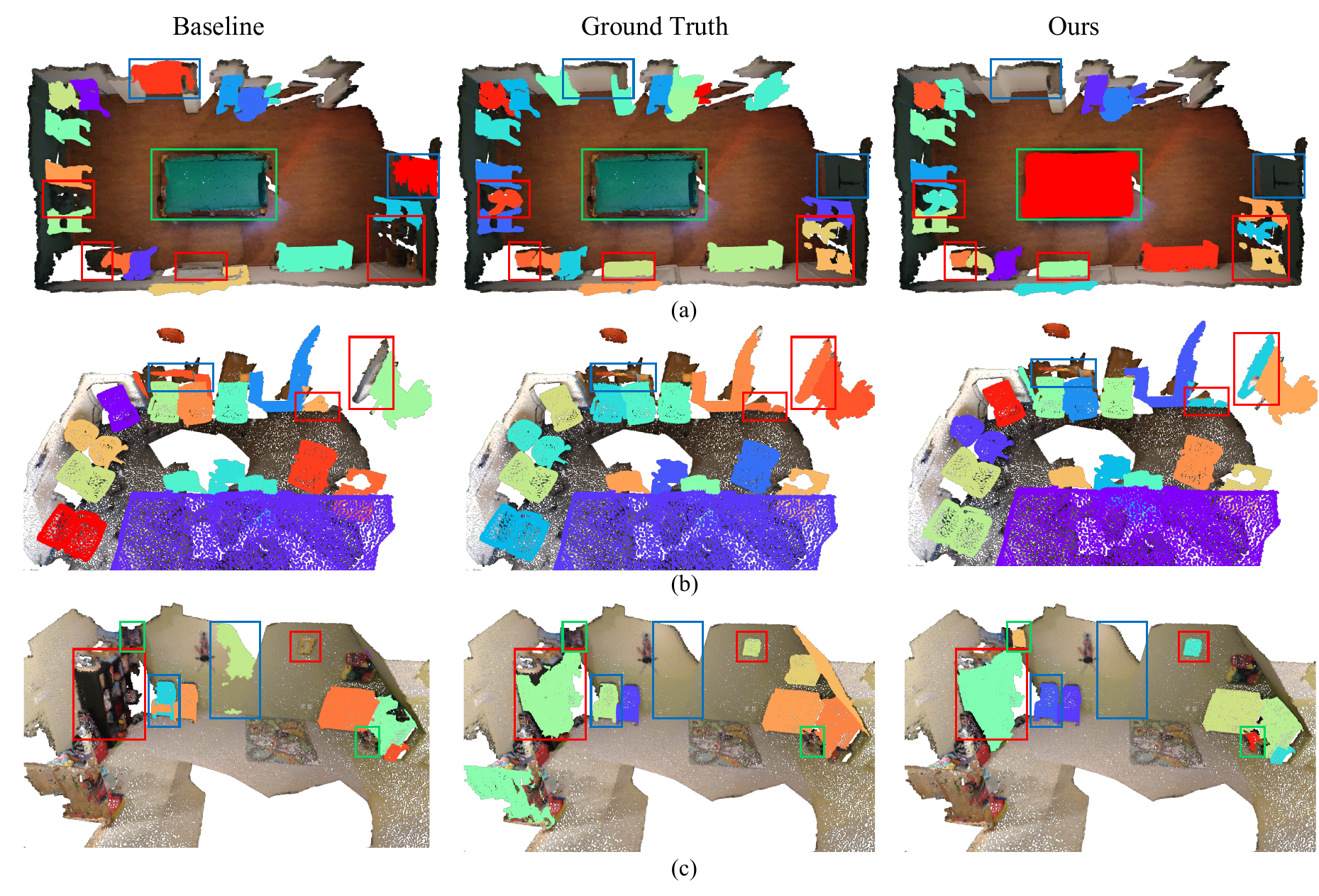}
    \caption{
        Visual comparison with the baseline method on the ScanNet200 validation set. 
        We visualize only the predictions with a confidence score greater than 0.5.
        The red boxes indicate instances missed by the baseline, the blue boxes indicate instances incorrectly segmented by the baseline, and the green boxes indicate additional instances segmented by our method.
    } 
    \label{fig.vis_comparison}
\end{figure*}

\noindent\textbf{3D Box Queries and Modulated Cross-Attention.} 
As described in Sec.\ref{sec:overview} and Sec.\ref{sec:prepare}, we introduce an additional positional component to the 3D object queries of the Transformer-based 3D instance segmentation model. 
Specifically, we denote it as $\mathbf{Q}^{P} \in \mathbb{R}^{M \times 6}$, where each query is formulated as an axis-aligned 3D bounding box $(x, y, z, l, w, h)$ in 3D space. 
Here, $(x, y, z)$ denotes the center coordinate of the box, and $(l, w, h)$ indicates its dimensions along the three axes.

We implement 3D box queries and modulated cross-attention in the following order. 
We first require the model to regress only the object center $(x, y, z)$, and then use the estimated center coordinates in the decoder to add positional embeddings for cross-attention. 
We apply supervision at each layer of the decoder and use the L1 loss.
In our implementation, we adopt a sine function as the positional encoding function $\operatorname{PE}()$, which maps each dimension of the 3D coordinate, i.e., $\{\mathbf{Q}^{P}_{[x]}, \mathbf{Q}^{P}_{[y]}, \mathbf{Q}^{P}_{[z]}\} \in \mathbb{R}^{M \times 1}$, into $\mathbb{R}^{M \times C'}$, where $C'=128$. 
Taking the $x$-coordinate as an example, the positional encoding function is formulated as:
\begin{equation}
    \begin{gathered}
        \operatorname{PE}(x)_{2i} = \sin\left(\frac{x}{T^{2i / C'}}\right), \\
        \operatorname{PE}(x)_{2i+1} = \cos\left(\frac{x}{T^{2i / C'}}\right),
    \end{gathered}
\end{equation}
where $T$ is a temperature hyperparameter that needs to be set manually, and the subscripts $2i$ and $2i+1$ denote the indices in the vector. 
Following DAB-DETR~\cite{liu2022dab} and DINO~\cite{zhang2022dino}, we set $T=20$.

In our implementation, we use the center coordinate of the superpoint corresponding to each 3D object query to initialize $(x, y, z)$, which is in metric scale. 
When updating via offset regression, the model is also required to predict metric values in the real 3D scene. 
However, before feeding the coordinates into the $\operatorname{PE}()$ function for positional encoding, we normalize them into the [0, 1] range.

Subsequently, we add a box size regression task to the model, enabling it to regress the full $(x, y, z, l, w, h)$ parameters. This allows our model to estimate axis-aligned 3D bounding boxes.
For the box size, we also apply layer-by-layer regression to predict the residual with respect to the ground truth. For initialization, we use a fixed value of $(l, w, h) = (0.5, 0.5, 0.5)$.

Finally, we use the estimated box size to modulate the positional attention map in the cross-attention, as detailed in Sec.~\ref{sec:bmca-3d} of the main paper.

We conduct an ablation study to verify the actual contribution of each component discussed above to the final performance, as shown in Sec.~\ref{ablation} of the main paper. 
The results demonstrate that significant improvements are only observed when we use box size to modulate the cross-attention, highlighting the effectiveness of the modulated cross-attention design.

\noindent\textbf{Cost Function and Loss Function.} 
We primarily follow the cost and loss function settings of SPFormer~\cite{spformer} and OneFormer3D~\cite{oneformer3d}, but on top of that, we additionally introduce a cost function and a loss function for regressing 3D bounding boxes.

For the cost function, we use the pair-wise matching cost $\mathcal{C}_{ik}$ to measure the similarity between the $i$-th prediction of the model and the $k$-th ground truth. It is computed as:
\begin{equation}
    \mathcal{C}_{ik}=-\beta_1\cdot p_{i,c_k}+\mathcal{C}_{ik}^{mask} + \beta_2 \cdot \mathcal{C}_{ik}^{box},
    \label{eq:cost_func}
\end{equation}
where $p_{i,c_k}$ represents the probability that the $i$-th prediction belongs to the semantic category $c_k$. $\beta_1$ is a weighting coefficient, which we set to $\beta_1 = 0.5$. 
$\mathcal{C}_{ik}^{mask}$ refers to the superpoint mask matching cost, which consists of a binary cross-entropy (BCE) loss and a Dice loss with Laplace smoothing~\cite{milletari2016v}. The detailed formula is:
\begin{equation}
    \mathcal{C}_{ik}^{mask}=\mathrm{BCE}(m_i,m_k^{gt})+1-2\frac{m_i\cdot m_k^{gt}+1}{|m_i|+\left|m_k^{gt}\right|+1},
\end{equation}
where $m_i$ and $m_k^{gt}$ denote the predicted and ground-truth superpoint masks, respectively.

In addition, we introduce a cost term $\mathcal{C}_{ik}^{box}$ to measure the accuracy of box regression, which is computed using the L1 distance function:
\begin{equation}
    \mathcal{C}_{ik}^{box} = \mathrm{L}_1(b_i, b_k^{gt})
\end{equation}
where $b_i$ and $b_k^{gt}$ denote the predicted and ground-truth 3D bounding boxes, respectively, both represented in the form of $(x, y, z, l, w, h)$. 
We set another weighting coefficient in Equation~\ref{eq:cost_func} as $\beta_2 = 0.5$.

For the loss function, as described in the implementation details in Sec.~\ref{sec:setup}, it can be formulated as:
\begin{equation}
    \mathcal{L} = \lambda_{1} \cdot \mathcal{L}_{cls} + \mathcal{L}_{bce} + \mathcal{L}_{dice} + \lambda_{2} \cdot \mathcal{L}_{sem} + \lambda_{3} \cdot \mathcal{L}_{box}
\end{equation}
where we set $\lambda_{1} = \lambda_{2} = \lambda_{3} = 0.5$ in our implementation. 
Among them, $\mathcal{L}_{cls}$, $\mathcal{L}_{bce}$, $\mathcal{L}_{dice}$, and $\mathcal{L}_{sem}$ are kept consistent with OneFormer3D, while $\mathcal{L}_{box}$ is an additional L1 loss.

\begin{figure*}[t]
    \centering
        \includegraphics[width=0.98\linewidth]{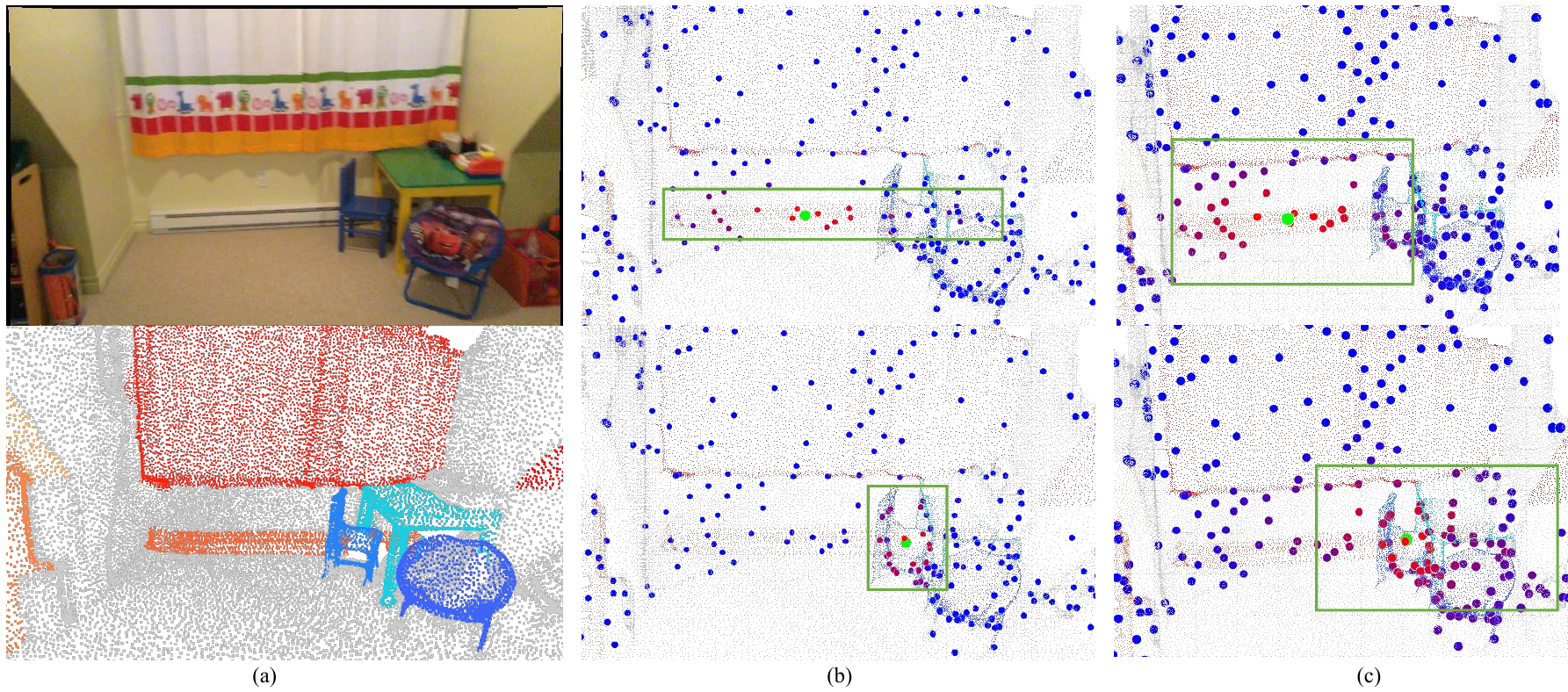}
    \caption{Visualization of the box-modulation's effect.
    (a) From top to bottom, we present an example RGB image from the input images and the ground‑truth segmentation results of this scene.
    (b) Shows the positional similarity within the box‑modulated cross‑attention module of our method. 
    The green point indicates the position of the current query, while the red and blue points represent the similarity between each superpoint and the query. 
    Red indicates high similarity, and blue indicates low similarity. 
    Regions with relatively high similarity are highlighted with green boxes. 
    Here, we present examples of a radiator and a chair, shown from top to bottom, respectively.
    (c) Depicts the positional similarity without applying box modulation.
    } 
    \label{fig.vis_modulation}
\end{figure*}

\subsection{Visualizations}
\label{sec_vis}

In this section, we present several visualization results to demonstrate the effectiveness of our method.

\noindent\textbf{Visual Comparison with Baseline.}
We present a visual comparison of segmentation results on the ScanNet200 validation set between {\methodname} and the baseline method (using the official checkpoint provided by OneFormer3D).
All instance proposals predicted by the models have confidence scores greater than 0.5.

As shown in Fig.~\ref{fig.vis_comparison}, red boxes indicate instances missed by the baseline, blue boxes denote instances incorrectly segmented by the baseline, and green boxes highlight additional instances segmented by our method.
In Fig.~\ref{fig.vis_comparison} (a), some tables, chairs, and a radiator are missed and not successfully segmented by the baseline. 
Moreover, the baseline incorrectly segmented the wall, which is not among the 198 instance categories in ScanNet200. 
In contrast, our method additionally segmented the table in the center of the scene, which was not annotated in the ground truth.

In Fig.~\ref{fig.vis_comparison} (b), the baseline failed to segment the window.

In Fig.~\ref{fig.vis_comparison} (c), the baseline missed the picture on the wall and the bookshelf on the left side of the room, and also produced incorrect segmentation for the wall and chairs. Furthermore, our method segmented two additional boxes, as indicated by the green boxes.

\noindent\textbf{Visualization of the Box-Modulation's Effect.} 
We visualize the positional similarity between 3D object queries and the 3D superpoints after and before the box modulation to show the effect of box modulation in our Box-Modulated Cross-Attention (BMCA-3D) module more intuitively. 

As shown in the top image of Fig.~\ref{fig.vis_modulation} (b), when box modulation is applied, the positional attention map corresponding to a query for a long radiator is modulated into a matching elongated shape. 
Similarly, in the bottom image, the attention map for a chair is adjusted according to the object’s specific dimensions.
In contrast, without box modulation, as shown in Fig.~\ref{fig.vis_modulation} (c), the positional map fails to adapt to the object’s size, which can further lead to incorrect information being retrieved during cross-attention.

In summary, after the box modulation, the similarity map transforms from an isotropic spherical shape to an ellipsoidal one that better aligns with the geometry of the target object, thereby guiding the attention to focus more accurately on points that belong to the object.

\begin{figure*}[hbt]
    \centering
        \includegraphics[width=0.98\linewidth]{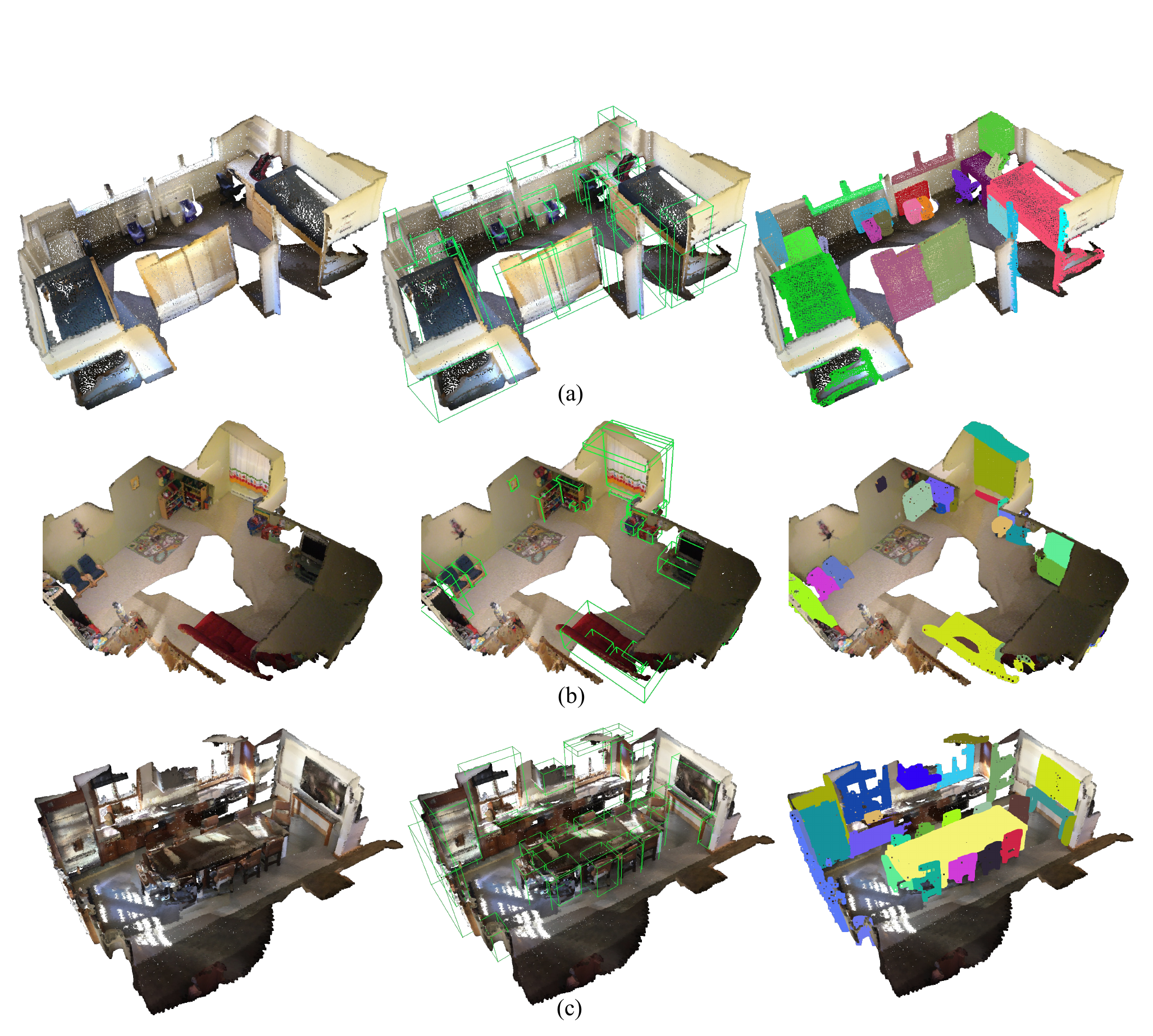}
    \caption{Visualization of the predicted 3D instance bounding boxes and 3D instance masks.
    } 
    \label{fig.vis_box_mask}
\end{figure*}

\noindent\textbf{Visualizations of Box and Mask Predictions.} 
As shown in Fig.~\ref{fig.vis_box_mask}, we show the qualitative predictions of each instance's 3D mask and the corresponding 3D bounding boxes.

\noindent\textbf{Visualizations of Segmentation Results In the Wild.} To evaluate the robustness of our method, we apply a reconstruction algorithm to in-the-wild videos and obtain noisy point clouds, which are then fed into our model for segmentation. Please refer to our supplementary materials for the videos, the reconstructed point clouds, and the segmentation results (\textit{Multimedia\_Appendix/wild\_video\_x directories}).

The results show that, although our model is only trained on the 1201 high-quality and clean scene point clouds from the ScanNetv2 dataset, it still exhibits strong robustness and generalization ability when handling noisy point clouds reconstructed from cluttered in-the-wild videos, demonstrating the superiority of our approach.

%% file: main.bbl
\begin{thebibliography}{52}
\providecommand{\natexlab}[1]{#1}

\bibitem[{Al~Khatib et~al.(2023)Al~Khatib, El~Amine~Boudjoghra, Lahoud, and Khan}]{al20233d}
Al~Khatib, S.; El~Amine~Boudjoghra, M.; Lahoud, J.; and Khan, F.~S. 2023.
\newblock {3D Instance Segmentation via Enhanced Spatial and Semantic Supervision}.
\newblock In \emph{Proceedings of the IEEE/CVF International Conference on Computer Vision}, 541--550.

\bibitem[{Arnaud et~al.(2025)Arnaud, McVay, Martin, Majumdar, Jatavallabhula, Thomas, Partsey, Dugas, Gejji, Sax et~al.}]{locate3d}
Arnaud, S.; McVay, P.; Martin, A.; Majumdar, A.; Jatavallabhula, K.~M.; Thomas, P.; Partsey, R.; Dugas, D.; Gejji, A.; Sax, A.; et~al. 2025.
\newblock {Locate 3D: Real-World Object Localization via Self-Supervised Learning in 3D}.
\newblock \emph{arXiv preprint arXiv:2504.14151}.

\bibitem[{Boudjoghra et~al.(2024)Boudjoghra, Dai, Lahoud, Cholakkal, Anwer, Khan, and Khan}]{openyolo3d}
Boudjoghra, M. E.~A.; Dai, A.; Lahoud, J.; Cholakkal, H.; Anwer, R.~M.; Khan, S.; and Khan, F.~S. 2024.
\newblock {Open-YOLO 3D: Towards Fast and Accurate Open-Vocabulary 3D Instance Segmentation}.
\newblock \emph{arXiv preprint arXiv:2406.02548}.

\bibitem[{Carion et~al.(2020)Carion, Massa, Synnaeve, Usunier, Kirillov, and Zagoruyko}]{detr}
Carion, N.; Massa, F.; Synnaeve, G.; Usunier, N.; Kirillov, A.; and Zagoruyko, S. 2020.
\newblock {End-to-End Object Detection with Transformers}.
\newblock In \emph{Computer Vision--ECCV 2020: 16th European Conference, Glasgow, UK, August 23--28, 2020, Proceedings, Part I 16}, 213--229. Springer.

\bibitem[{Chen et~al.(2021)Chen, Fang, Zhang, Liu, and Wang}]{chen2021hierarchical}
Chen, S.; Fang, J.; Zhang, Q.; Liu, W.; and Wang, X. 2021.
\newblock {Hierarchical Aggregation for 3D Instance Segmentation}.
\newblock In \emph{Proceedings of the IEEE/CVF International Conference on Computer Vision}, 15467--15476.

\bibitem[{Choy, Gwak, and Savarese(2019)}]{minkowski}
Choy, C.; Gwak, J.; and Savarese, S. 2019.
\newblock {4D Spatio-Temporal ConvNets: Minkowski Convolutional Neural Networks}.
\newblock In \emph{Proceedings of the IEEE/CVF conference on computer vision and pattern recognition}, 3075--3084.

\bibitem[{Dai et~al.(2017)Dai, Chang, Savva, Halber, Funkhouser, and Nie{\ss}ner}]{scannet}
Dai, A.; Chang, A.~X.; Savva, M.; Halber, M.; Funkhouser, T.; and Nie{\ss}ner, M. 2017.
\newblock {ScanNet: Richly-Annotated 3D Reconstructions of Indoor Scenes}.
\newblock In \emph{Proceedings of the IEEE conference on computer vision and pattern recognition}, 5828--5839.

\bibitem[{Engelmann et~al.(2020)Engelmann, Bokeloh, Fathi, Leibe, and Nie{\ss}ner}]{engelmann20203d}
Engelmann, F.; Bokeloh, M.; Fathi, A.; Leibe, B.; and Nie{\ss}ner, M. 2020.
\newblock {3D-MPA: Multi-Proposal Aggregation for 3D Semantic Instance Segmentation}.
\newblock In \emph{Proceedings of the IEEE/CVF conference on computer vision and pattern recognition}, 9031--9040.

\bibitem[{Gu et~al.(2024)Gu, Kuwajerwala, Morin, Jatavallabhula, Sen, Agarwal, Rivera, Paul, Ellis, Chellappa et~al.}]{gu2024conceptgraphs}
Gu, Q.; Kuwajerwala, A.; Morin, S.; Jatavallabhula, K.~M.; Sen, B.; Agarwal, A.; Rivera, C.; Paul, W.; Ellis, K.; Chellappa, R.; et~al. 2024.
\newblock {ConceptGraphs: Open-Vocabulary 3D Scene Graphs for Perception and Planning}.
\newblock In \emph{2024 IEEE International Conference on Robotics and Automation (ICRA)}, 5021--5028. IEEE.

\bibitem[{Hou, Dai, and Nie{\ss}ner(2019)}]{hou20193d}
Hou, J.; Dai, A.; and Nie{\ss}ner, M. 2019.
\newblock {3D-SIS: 3D Semantic Instance Segmentation of RGB-D Scans}.
\newblock In \emph{Proceedings of the IEEE/CVF conference on computer vision and pattern recognition}, 4421--4430.

\bibitem[{Jain et~al.(2024)Jain, Katara, Gkanatsios, Harley, Sarch, Aggarwal, Chaudhary, and Fragkiadaki}]{ODIN}
Jain, A.; Katara, P.; Gkanatsios, N.; Harley, A.~W.; Sarch, G.; Aggarwal, K.; Chaudhary, V.; and Fragkiadaki, K. 2024.
\newblock {ODIN: A Single Model for 2D and 3D Segmentation}.
\newblock In \emph{Proceedings of the IEEE/CVF Conference on Computer Vision and Pattern Recognition}, 3564--3574.

\bibitem[{Jiang et~al.(2020{\natexlab{a}})Jiang, Yan, Cai, Zheng, and Xiao}]{jiang2020end}
Jiang, H.; Yan, F.; Cai, J.; Zheng, J.; and Xiao, J. 2020{\natexlab{a}}.
\newblock {End-to-End 3D Point Cloud Instance Segmentation Without Detection}.
\newblock In \emph{Proceedings of the IEEE/CVF Conference on Computer Vision and Pattern Recognition}, 12796--12805.

\bibitem[{Jiang et~al.(2020{\natexlab{b}})Jiang, Zhao, Shi, Liu, Fu, and Jia}]{jiang2020pointgroup}
Jiang, L.; Zhao, H.; Shi, S.; Liu, S.; Fu, C.-W.; and Jia, J. 2020{\natexlab{b}}.
\newblock {PointGroup: Dual-Set Point Grouping for 3D Instance Segmentation}.
\newblock In \emph{Proceedings of the IEEE/CVF conference on computer vision and Pattern recognition}, 4867--4876.

\bibitem[{Kaufman and Rousseeuw(2009)}]{kaufman2009finding}
Kaufman, L.; and Rousseeuw, P.~J. 2009.
\newblock \emph{Finding groups in data: an introduction to cluster analysis}.
\newblock John Wiley \& Sons.

\bibitem[{Kirillov et~al.(2023)Kirillov, Mintun, Ravi, Mao, Rolland, Gustafson, Xiao, Whitehead, Berg, Lo et~al.}]{SAM}
Kirillov, A.; Mintun, E.; Ravi, N.; Mao, H.; Rolland, C.; Gustafson, L.; Xiao, T.; Whitehead, S.; Berg, A.~C.; Lo, W.-Y.; et~al. 2023.
\newblock {Segment Anything}.
\newblock In \emph{Proceedings of the IEEE/CVF international conference on computer vision}, 4015--4026.

\bibitem[{Kolodiazhnyi et~al.(2024{\natexlab{a}})Kolodiazhnyi, Vorontsova, Konushin, and Rukhovich}]{oneformer3d}
Kolodiazhnyi, M.; Vorontsova, A.; Konushin, A.; and Rukhovich, D. 2024{\natexlab{a}}.
\newblock {OneFormer3D: One Transformer for Unified Point Cloud Segmentation}.
\newblock In \emph{Proceedings of the IEEE/CVF Conference on Computer Vision and Pattern Recognition}, 20943--20953.

\bibitem[{Kolodiazhnyi et~al.(2024{\natexlab{b}})Kolodiazhnyi, Vorontsova, Konushin, and Rukhovich}]{td3d}
Kolodiazhnyi, M.; Vorontsova, A.; Konushin, A.; and Rukhovich, D. 2024{\natexlab{b}}.
\newblock {Top-Down Beats Bottom-Up in 3D Instance Segmentation}.
\newblock In \emph{Proceedings of the IEEE/CVF Winter Conference on Applications of Computer Vision}, 3566--3574.

\bibitem[{Lai et~al.(2023)Lai, Yuan, Chu, Chen, Hu, and Jia}]{maft}
Lai, X.; Yuan, Y.; Chu, R.; Chen, Y.; Hu, H.; and Jia, J. 2023.
\newblock {Mask-Attention-Free Transformer for 3D Instance Segmentation}.
\newblock In \emph{Proceedings of the IEEE/CVF International Conference on Computer Vision}, 3693--3703.

\bibitem[{Landrieu and Simonovsky(2018)}]{landrieu2018large}
Landrieu, L.; and Simonovsky, M. 2018.
\newblock {Large-Scale Point Cloud Semantic Segmentation With Superpoint Graphs}.
\newblock In \emph{Proceedings of the IEEE conference on computer vision and pattern recognition}, 4558--4567.

\bibitem[{Li et~al.(2022)Li, Wang, Li, Xie, Sima, Lu, Qiao, and Dai}]{li2022bevformer}
Li, Z.; Wang, W.; Li, H.; Xie, E.; Sima, C.; Lu, T.; Qiao, Y.; and Dai, J. 2022.
\newblock {BEVFormer: Learning Bird's-Eye-View Representation from Multi-Camera Images via Spatiotemporal Transformers}.
\newblock In \emph{Computer Vision--ECCV 2022: 17th European Conference, Tel Aviv, Israel, October 23--27, 2022, Proceedings, Part IX}, 1--18. Springer.

\bibitem[{Liang et~al.(2021)Liang, Li, Xu, Tan, and Jia}]{liang2021instance}
Liang, Z.; Li, Z.; Xu, S.; Tan, M.; and Jia, K. 2021.
\newblock {Instance Segmentation in 3D Scenes Using Semantic Superpoint Tree Networks}.
\newblock In \emph{Proceedings of the IEEE/CVF international conference on computer vision}, 2783--2792.

\bibitem[{Liu et~al.(2021)Liu, Li, Zhang, Yang, Qi, Su, Zhu, and Zhang}]{liu2022dab}
Liu, S.; Li, F.; Zhang, H.; Yang, X.; Qi, X.; Su, H.; Zhu, J.; and Zhang, L. 2021.
\newblock {DAB-DETR: Dynamic Anchor Boxes are Better Queries for DETR}.
\newblock In \emph{International Conference on Learning Representations}.

\bibitem[{Liu et~al.(2024)Liu, Zeng, Ren, Li, Zhang, Yang, Jiang, Li, Yang, Su et~al.}]{groundingdino}
Liu, S.; Zeng, Z.; Ren, T.; Li, F.; Zhang, H.; Yang, J.; Jiang, Q.; Li, C.; Yang, J.; Su, H.; et~al. 2024.
\newblock {Grounding DINO: Marrying DINO with Grounded Pre-training for Open-Set Object Detection}.
\newblock In \emph{European Conference on Computer Vision}, 38--55. Springer.

\bibitem[{Loshchilov(2017)}]{loshchilov2017decoupled}
Loshchilov, I. 2017.
\newblock {Decoupled Weight Decay Regularization}.
\newblock \emph{arXiv preprint arXiv:1711.05101}.

\bibitem[{Lu and Deng(2025)}]{lu2025relation3d}
Lu, J.; and Deng, J. 2025.
\newblock {Relation3D: Enhancing Relation Modeling for Point Cloud Instance Segmentation}.
\newblock In \emph{Proceedings of the Computer Vision and Pattern Recognition Conference}, 8889--8899.

\bibitem[{Lu et~al.(2023)Lu, Deng, Wang, He, and Zhang}]{queryformer}
Lu, J.; Deng, J.; Wang, C.; He, J.; and Zhang, T. 2023.
\newblock {Query Refinement Transformer for 3D Instance Segmentation}.
\newblock In \emph{Proceedings of the IEEE/CVF International Conference on Computer Vision}, 18516--18526.

\bibitem[{Milletari, Navab, and Ahmadi(2016)}]{milletari2016v}
Milletari, F.; Navab, N.; and Ahmadi, S.-A. 2016.
\newblock {V-Net: Fully Convolutional Neural Networks for Volumetric Medical Image Segmentation}.
\newblock In \emph{2016 fourth international conference on 3D vision (3DV)}, 565--571. Ieee.

\bibitem[{Mur-Artal and Tard{\'o}s(2017)}]{mur2017orb}
Mur-Artal, R.; and Tard{\'o}s, J.~D. 2017.
\newblock {ORB-SLAM2: An Open-Source SLAM System for Monocular, Stereo, and RGB-D Cameras}.
\newblock \emph{IEEE transactions on robotics}, 33(5): 1255--1262.

\bibitem[{Nguyen et~al.(2024)Nguyen, Ngo, Kalogerakis, Gan, Tran, Pham, and Nguyen}]{open3dis}
Nguyen, P.; Ngo, T.~D.; Kalogerakis, E.; Gan, C.; Tran, A.; Pham, C.; and Nguyen, K. 2024.
\newblock {Open3DIS: Open-Vocabulary 3D Instance Segmentation with 2D Mask Guidance}.
\newblock In \emph{Proceedings of the IEEE/CVF Conference on Computer Vision and Pattern Recognition}, 4018--4028.

\bibitem[{Oquab et~al.(2023)Oquab, Darcet, Moutakanni, Vo, Szafraniec, Khalidov, Fernandez, Haziza, Massa, El-Nouby et~al.}]{DINOv2}
Oquab, M.; Darcet, T.; Moutakanni, T.; Vo, H.; Szafraniec, M.; Khalidov, V.; Fernandez, P.; Haziza, D.; Massa, F.; El-Nouby, A.; et~al. 2023.
\newblock {DINOv2: Learning Robust Visual Features without Supervision}.
\newblock \emph{arXiv preprint arXiv:2304.07193}.

\bibitem[{Peng et~al.(2023)Peng, Genova, Jiang, Tagliasacchi, Pollefeys, Funkhouser et~al.}]{openscene}
Peng, S.; Genova, K.; Jiang, C.; Tagliasacchi, A.; Pollefeys, M.; Funkhouser, T.; et~al. 2023.
\newblock {OpenScene: 3D Scene Understanding With Open Vocabularies}.
\newblock In \emph{Proceedings of the IEEE/CVF conference on computer vision and pattern recognition}, 815--824.

\bibitem[{Radford et~al.(2021)Radford, Kim, Hallacy, Ramesh, Goh, Agarwal, Sastry, Askell, Mishkin, Clark et~al.}]{clip}
Radford, A.; Kim, J.~W.; Hallacy, C.; Ramesh, A.; Goh, G.; Agarwal, S.; Sastry, G.; Askell, A.; Mishkin, P.; Clark, J.; et~al. 2021.
\newblock {Learning Transferable Visual Models From Natural Language Supervision}.
\newblock In \emph{International conference on machine learning}, 8748--8763. PmLR.

\bibitem[{Ren et~al.(2024)Ren, Chen, Jiang, Zeng, Xiong, Liu, Ma, Shen, Gao, Jiang et~al.}]{DINOX}
Ren, T.; Chen, Y.; Jiang, Q.; Zeng, Z.; Xiong, Y.; Liu, W.; Ma, Z.; Shen, J.; Gao, Y.; Jiang, X.; et~al. 2024.
\newblock {DINO-X: A Unified Vision Model for Open-World Object Detection and Understanding}.
\newblock \emph{arXiv preprint arXiv:2411.14347}.

\bibitem[{Rozenberszki, Litany, and Dai(2022)}]{scannet200}
Rozenberszki, D.; Litany, O.; and Dai, A. 2022.
\newblock {Language-Grounded Indoor 3D Semantic Segmentation in the Wild}.
\newblock In \emph{European Conference on Computer Vision}, 125--141. Springer.

\bibitem[{Sch\"{o}nberger et~al.(2016)Sch\"{o}nberger, Zheng, Pollefeys, and Frahm}]{schoenberger2016mvs}
Sch\"{o}nberger, J.~L.; Zheng, E.; Pollefeys, M.; and Frahm, J.-M. 2016.
\newblock {Pixelwise View Selection for Unstructured Multi-View Stereo}.
\newblock In \emph{European Conference on Computer Vision (ECCV)}.

\bibitem[{Schult et~al.(2023)Schult, Engelmann, Hermans, Litany, Tang, and Leibe}]{Mask3D}
Schult, J.; Engelmann, F.; Hermans, A.; Litany, O.; Tang, S.; and Leibe, B. 2023.
\newblock {Mask3D: Mask Transformer for 3D Semantic Instance Segmentation}.
\newblock In \emph{2023 IEEE International Conference on Robotics and Automation (ICRA)}, 8216--8223. IEEE.

\bibitem[{Shin et~al.(2024)Shin, Zhou, Vankadari, Markham, and Trigoni}]{shin2024spherical}
Shin, S.; Zhou, K.; Vankadari, M.; Markham, A.; and Trigoni, N. 2024.
\newblock {Spherical Mask: Coarse-to-Fine 3D Point Cloud Instance Segmentation with Spherical Representation}.
\newblock In \emph{Proceedings of the IEEE/CVF Conference on Computer Vision and Pattern Recognition}, 4060--4069.

\bibitem[{Sun et~al.(2023)Sun, Qing, Tan, and Xu}]{spformer}
Sun, J.; Qing, C.; Tan, J.; and Xu, X. 2023.
\newblock {Superpoint Transformer for 3D Scene Instance Segmentation}.
\newblock In \emph{Proceedings of the AAAI Conference on Artificial Intelligence}, volume~37, 2393--2401.

\bibitem[{Takmaz et~al.(2023)Takmaz, Fedele, Sumner, Pollefeys, Tombari, and Engelmann}]{openmask3d}
Takmaz, A.; Fedele, E.; Sumner, R.~W.; Pollefeys, M.; Tombari, F.; and Engelmann, F. 2023.
\newblock {OpenMask3D: Open-Vocabulary 3D Instance Segmentation}.
\newblock \emph{arXiv preprint arXiv:2306.13631}.

\bibitem[{Vu et~al.(2022)Vu, Kim, Luu, Nguyen, and Yoo}]{vu2022softgroup}
Vu, T.; Kim, K.; Luu, T.~M.; Nguyen, T.; and Yoo, C.~D. 2022.
\newblock {SoftGroup for 3D Instance Segmentation on Point Clouds}.
\newblock In \emph{Proceedings of the IEEE/CVF conference on computer vision and pattern recognition}, 2708--2717.

\bibitem[{Wang et~al.(2024{\natexlab{a}})Wang, Liu, Gong, Quan, and Wang}]{competitorformer}
Wang, D.; Liu, J.; Gong, H.; Quan, Y.; and Wang, D. 2024{\natexlab{a}}.
\newblock {CompetitorFormer: Competitor Transformer for 3D Instance Segmentation}.
\newblock \emph{arXiv preprint arXiv:2411.14179}.

\bibitem[{Wang et~al.(2025)Wang, Chen, Karaev, Vedaldi, Rupprecht, and Novotny}]{vggt}
Wang, J.; Chen, M.; Karaev, N.; Vedaldi, A.; Rupprecht, C.; and Novotny, D. 2025.
\newblock {VGGT: Visual Geometry Grounded Transformer}.
\newblock In \emph{Proceedings of the Computer Vision and Pattern Recognition Conference}, 5294--5306.

\bibitem[{Wang et~al.(2024{\natexlab{b}})Wang, Leroy, Cabon, Chidlovskii, and Revaud}]{dust3r}
Wang, S.; Leroy, V.; Cabon, Y.; Chidlovskii, B.; and Revaud, J. 2024{\natexlab{b}}.
\newblock {DUSt3R: Geometric 3D Vision Made Easy}.
\newblock In \emph{Proceedings of the IEEE/CVF Conference on Computer Vision and Pattern Recognition}, 20697--20709.

\bibitem[{Wang et~al.(2019)Wang, Liu, Shen, Shen, and Jia}]{wang2019associatively}
Wang, X.; Liu, S.; Shen, X.; Shen, C.; and Jia, J. 2019.
\newblock {Associatively Segmenting Instances and Semantics in Point Clouds}.
\newblock In \emph{Proceedings of the IEEE/CVF conference on computer vision and pattern recognition}, 4096--4105.

\bibitem[{Xie et~al.(2021)Xie, Xiang, Mousavian, and Fox}]{xie2021unseen}
Xie, C.; Xiang, Y.; Mousavian, A.; and Fox, D. 2021.
\newblock {Unseen Object Instance Segmentation for Robotic Environments}.
\newblock \emph{IEEE Transactions on Robotics}, 37(5): 1343--1359.

\bibitem[{Yang et~al.(2019)Yang, Wang, Clark, Hu, Wang, Markham, and Trigoni}]{yang2019learning}
Yang, B.; Wang, J.; Clark, R.; Hu, Q.; Wang, S.; Markham, A.; and Trigoni, N. 2019.
\newblock {Learning Object Bounding Boxes for 3D Instance Segmentation on Point Clouds}.
\newblock \emph{Advances in neural information processing systems}, 32.

\bibitem[{Yang et~al.(2023)Yang, Wu, He, Zhao, and Liu}]{SAM3D}
Yang, Y.; Wu, X.; He, T.; Zhao, H.; and Liu, X. 2023.
\newblock {SAM3D: Segment Anything in 3D Scenes}.
\newblock \emph{arXiv preprint arXiv:2306.03908}.

\bibitem[{Yi et~al.(2019)Yi, Zhao, Wang, Sung, and Guibas}]{yi2019gspn}
Yi, L.; Zhao, W.; Wang, H.; Sung, M.; and Guibas, L.~J. 2019.
\newblock {GSPN: Generative Shape Proposal Network for 3D Instance Segmentation in Point Cloud}.
\newblock In \emph{Proceedings of the IEEE/CVF conference on computer vision and pattern recognition}, 3947--3956.

\bibitem[{Yin et~al.(2024)Yin, Liu, Xiao, Cohen-Or, Huang, and Chen}]{SAI3D}
Yin, Y.; Liu, Y.; Xiao, Y.; Cohen-Or, D.; Huang, J.; and Chen, B. 2024.
\newblock {SAI3D: Segment Any Instance in 3D Scenes}.
\newblock In \emph{Proceedings of the IEEE/CVF Conference on Computer Vision and Pattern Recognition}, 3292--3302.

\bibitem[{Zeid et~al.(2025)Zeid, Yilmaz, de~Geus, Hermans, Adrian, Linder, and Leibe}]{DITR}
Zeid, K.~A.; Yilmaz, K.; de~Geus, D.; Hermans, A.; Adrian, D.; Linder, T.; and Leibe, B. 2025.
\newblock {DINO in the Room: Leveraging 2D Foundation Models for 3D Segmentation}.
\newblock \emph{arXiv preprint arXiv:2503.18944}.

\bibitem[{Zhang and Wonka(2021)}]{zhang2021point}
Zhang, B.; and Wonka, P. 2021.
\newblock {Point Cloud Instance Segmentation Using Probabilistic Embeddings}.
\newblock In \emph{Proceedings of the IEEE/CVF Conference on Computer Vision and Pattern Recognition}, 8883--8892.

\bibitem[{Zhang et~al.(2022)Zhang, Li, Liu, Zhang, Su, Zhu, Ni, and Shum}]{zhang2022dino}
Zhang, H.; Li, F.; Liu, S.; Zhang, L.; Su, H.; Zhu, J.; Ni, L.; and Shum, H.-Y. 2022.
\newblock {DINO: DETR with Improved DeNoising Anchor Boxes for End-to-End Object Detection}.
\newblock In \emph{The Eleventh International Conference on Learning Representations}.

\end{thebibliography}
